\documentclass[pdflatex,sn-mathphys-num]{sn-jnl}

\usepackage{graphicx}%
\usepackage{multirow}%
\usepackage{amsmath,amssymb,amsfonts}%
\usepackage{amsthm}%
\usepackage{mathrsfs}%
\usepackage[title]{appendix}%
\usepackage{xcolor}%
\usepackage{textcomp}%
\usepackage{manyfoot}%
\usepackage{booktabs}%
\usepackage{algorithm}%
\usepackage{algorithmicx}%
\usepackage{algpseudocode}%
\usepackage{listings}%

\theoremstyle{thmstyleone}%
\theoremstyle{thmstyletwo}%

\theoremstyle{thmstylethree}%

\begin{document}

\title[Article Title]{NepEMO: A Multi-Label Emotion and Sentiment Analysis on Nepali Reddit with Linguistic Insights and Temporal Trends}


\author *[1]{\fnm{Sameer} \sur{Sitoula}} \email{sameersitaula67@gmail.com}

\author [2]{\fnm{Tej Bahadur} \sur{Shahi}} \email{t.shahi@qut.edu.au}

\author[3]{\fnm{Laxmi Prasad} \sur{Bhatt}} \email{lpbhatta0828@gmail.com}

\author[3]{\fnm{Anisha} \sur{Pokhrel}} \email{anishapokhrel01@gmail.com}

\author[4]{\fnm{Arjun} \sur{Neupane}} \email{a.neupane@cqu.edu.au}

\affil *[1]{\orgdiv{Department of Computer Science and Application}, \orgname{Mechi Multiple Campus, Tribhuvan University}, \orgaddress{ \city{Bhadrapur}, \postcode{57200}, \state{Koshi}, \country{Nepal}}}

\affil [2]{\orgdiv{School of Computer Science}, \orgname{Queensland University of Technology}, \orgaddress{\street{George Street}, \city{Brisbane}, \postcode{4000}, \state{QLD}, \country{Australia}}}

\affil[3]{\orgdiv{Department of Electronics and Computer Engineering}, \orgname{Advanced College of Engineering and Management, Tribhuvan University}, \orgaddress{\street{Kalanki}, \city{Kathmandu}, \postcode{44600}, \state{Bagmati}, \country{Nepal}}}

\affil[4]{\orgdiv{School of Engineering and Technology}, \orgname{Central Queensland University}, \orgaddress{\street{Norman Gardens}, \city{Rockhampton}, \postcode{4701}, \state{QLD}, \country{Australia}}}


\abstract{Social media (SM) platforms (e.g. Facebook, Twitter, and Reddit) are increasingly leveraged to share opinions and emotions, specifically during challenging events, such as natural disasters, pandemics, and political elections, and joyful occasions like festivals and celebrations. Among the SM platforms, Reddit provides a unique space for its users to anonymously express their experiences and thoughts on sensitive issues such as health and daily life. In this work, we present a novel dataset, called NepEMO, for multi-label emotion (MLE) and sentiment classification (SC) on the Nepali subreddit post. We curate and build a manually annotated dataset of 4,462 posts (January 2019- June 2025) written in English, Romanised Nepali and Devanagari script for five emotions (fear, anger, sadness, joy, and depression) and three sentiment classes (positive, negative, and neutral). 
We perform a detailed analysis of posts to capture linguistic insights, including emotion trends, co-occurrence of emotions, sentiment-specific n-grams, and topic modelling using Latent Dirichlet Allocation and TF-IDF keyword extraction. Finally, we compare various traditional machine learning (ML), deep learning (DL), and transformer models for MLE and SC tasks. The result shows that transformer models consistently outperform the ML and DL models for both tasks. 
}




\maketitle
\section{Introduction}
\label{introduction}
 In recent times, people have shared their opinions, emotions, and experiences on various social media (SM) platforms. Whenever they are seeking help or want to share their thoughts, they log into social networks such as Twitter, Facebook, or Reddit \cite{dolan2019social}. In return, they receive others' responses in the form of comments \cite{choi2014social}. These social networks are used more and more during challenging times, such as the COVID-19 pandemic \cite{tsao2021social}, natural disasters \cite{kongthon2014role}, and many other occasions, such as political elections \cite{thapa2023nehate}. Platforms like Reddit provide a favourable environment for people to participate in discourse ranging from daily life to emotionally challenging topics \cite{de2014mental}.  Because of its anonymous nature and user privacy preservation model, Reddit has become a popular platform for users to openly share their personal thoughts on various topics such as politics, social issues, and mental health. It provides various groups known as subreddits where users can ask for recommendations and guidance on broad topics \cite{proferes2021studying}. Reddit also has a Nepali subreddit where Nepali people from all over the world can express themselves. Users often post in English or code-mixed (e.g., a mix of English, Roman Nepali or even Devanagari script), often expressing complex emotional experiences that reflect Nepal's social and psychological realities.
 
 Extracting insightful information from these SM posts can help identify people's emotions and their mental health status during difficult times \cite{zad2021emotion}. Therefore, the automated processing of such text has become essential for analysing and understanding the multilingual and emotionally rich expressions found on SM platforms like Reddit, making it a crucial area within natural language processing (NLP) \cite{tunca2025algorithms}. Human emotions are often more intricate and complex. For example, someone who talks about their last examination might feel both fear and joy at the same time. Or, a person remembering their loved one might feel joy and sadness together. To capture such mixed human emotion, the traditional single-label emotion (SLE) classification methods need to be extended to multi-label emotion (MLE) classification. 
In MLE classification, each SM text is associated with multiple emotion labels simultaneously. For example, a code-mixed SM post \textit {`I’m nervous but excited for my new job — man ta dherai khusi cha tara alik dar pani lagira cha.'} translated as \textit {`I’m nervous but excited for my new job — my heart feels very happy but I’m also a little scared.'} convey the emotion as Fear \& Joy \cite{zhang2013review, liu2017deep}. SLE models often overlook such mixed emotions, but MLE classification can capture these nuances more accurately. One prominent dataset in this field is Google's GoEmotions \cite{demszky2020goemotions}, which contains over 58,000 Reddit comments annotated for 27 distinct emotional categories \cite{demszky2020goemotions} 

Despite the significant advances in MLE for high-resource languages such as English, Spanish, or Mandarin \cite{demszky2020goemotions,nandwani2021review,buechel2022emobank}, this has not been explored for emotion analysis in Nepali text. Most of the existing works in Nepali SM text analysis have focused on a single-label SC task \cite{sitaula2021deep,sitoula2025sentiment}, which classifies the SM text either into positive, negative or neutral. Furthermore, Nepali SM text is often characterised by the use of mixed scripts, with users combining English, Roman Nepali, and Devanagari in a single post. This code-mixing phenomenon presents a significant challenge for existing emotion detection models, which typically do not account for mixed-language text. 

Nepali is one of the major and official languages of Nepal and is spoken by more than 40 million people worldwide \cite{rajan2022survey,khatiwada_2009_nepali,khadka2019multilingual,timilsina2022nepberta}. And usually Nepali-originating individual express their emotions in SM not only in English but also in Nepali script. Recently, Nepal has experienced a multitude of crises, such as the 2015 earthquake, the global COVID-19 pandemic, political instability and the Zen-Z protest (The protest was against corruption, unemployment, and a proposed social media ban, and more than 70 people were killed, with hundreds injured, ultimately resulting in the collapse of the government in Nepal \footnote{\url{https://www.bbc.com/news/articles/crkj0lzlr3ro}, accessed on September 18, 2025}). Due to these events, Nepalese have been deeply affected by stress and anxiety, fear, sadness, and depression \cite{kane2018mental}. For example, many people faced severe economic hardship due to job losses during the COVID-19 pandemic \cite{sigdel2020depression}, which exacerbated feelings of despair and uncertainty. Additionally, political instability and corruption \cite{zipperer2025digging} have led to economic stagnation and rising unemployment, further straining the mental well-being of the people.  

In this study, we explore the trend and evolution of mental health discussions and emotional expressions reflected by Nepalese online communities on Reddit. To achieve this, we develop the NepEMO dataset of 4,462 posts from selected Nepali subreddits using keywords related to mental health and well-being, covering five emotion labels: fear, anger, sadness, depression, joy, and three sentiment categories: positive, negative, and neutral. We analyse differences and similarities in emotional expression, particularly contrasting negative emotions such as anger, fear, sadness, and depression with the positive emotion of joy. Furthermore, we examine temporal trends to observe how specific emotions evolve, investigate emotional co-occurrence, and identify commonly used bigrams and trigrams for each sentiment. Lastly, we benchmark the performance of various machine learning  (ML) and deep learning  (DL) models on emotion and sentiment classification tasks. 


In summary, our paper has the following contributions:

\begin{itemize}
    \item [(i)] We introduce a Novel NepEMO dataset containing code-mixed text (English, Roman-Nepali, and Devanagari) labelled with five emotional categories (fear, anger, sadness, joy, and depression) and three sentiment classes (positive, negative, and neutral).
    \item [(ii)] We present a comprehensive linguistic and temporal analysis to understand the evolution and trends in emotional expression over six years (January 2019 to June 2025 ).
    \item [(iii)] We explore co-occurrence patterns of emotional keywords and frequently used sentiment-specific phrases, offering deep insights into Nepali online discourse on mental health and well-being.
    \item [(iv)] We benchmark several traditional machine learning (ML) and recent deep learning (DL) models, demonstrating their performance on emotion and sentiment classification tasks on code-mixed Nepali datasets.
\end{itemize}

The paper is organised as follows. Section \ref{related work} contains related work on MLE, SC, and mental health NLP in English, code-mixed, and Nepali languages. Section \ref{dataset} presents our data collection and annotation process. Section \ref{methods} contains the various techniques used in our work, such as how bag-of-words (BoW)-based features are used, an explanation of ML, DL, and multilingual transformer-based models, followed by implementation details and performance metrics. Section \ref{result_and_discussion} contains the results and discussions of ML, DL, and transformer-based models. Finally, Section \ref{conclusion} concludes the paper with future work and limitations.

\section{Related Work}
\label{related work}
MLE classifies the text into multiple labels (e.g., joy, anger, sadness, depression), whereas the SC classifies the opinions into positive, negative or neutral classes. Emotion and sentiment analysis are well-studied in high-resource language settings such as English and Spanish \cite{nandwani2021review,buechel2022emobank} that utilise various techniques for automated emotion and sentiment classification, including  ML, DL and large language models (LLM). However, a few works have been done for Nepali sentiment analysis, whereas there is not much work that exists for MLE analysis in the Nepali language. Therefore, we review the overall literature related to emotion and sentiment analysis in both Nepali and non-Nepali languages.

Traditional ML approaches, such as support vector machine (SVM), Decision tree (DT), and Naive Bayes (NB), have been widely explored for SC and emotion detection tasks. For example, Saad et al. \cite{saad2018evaluation} compared the DT and SVM models for emotion detection, reporting that SVM achieved higher accuracy, which established an initial benchmark for emotion classification. A neural network algorithm was proposed by Zhang et al. \cite{zhang2006multilabel} for multi-label text categorisation. Their method uses a novel error function that trains the network to give higher scores to the correct labels and lower scores to the incorrect ones. Their research demonstrated that their method is effective in tasks such as text categorisation and gene analysis. This work is significant as it adapts neural networks to solve multi-label classification problems more accurately. Later, Tsoumakas and Katakis et al. \cite{tsoumakas2008multi} provided comprehensive surveys on multi-label classification. They introduced a way to describe a multi-label dataset, which has label cardinality and label density. Their work covers datasets from broad domains such as bioinformatics and image annotation. This study lays the foundation for synthesising many ideas, marking a significant starting point for future research in this field.

With the rise of DL, transformer-based architectures have become the foundation for powerful multilingual pre-trained models, such as multilingual Bidirectional encoder representations from transformers (mBERT)\cite{gardazi2025bert}, and XLM-RoBERTa \cite{sitoula2025sentiment}. These models were found to outperform the monolingual baseline models in the SC and MLE tasks. For instance, Ranasinghe and Zampierie et al. \cite{ranasinghe2020multilingual} fine-tuned an XLM-RoBERTa model for languages such as Bengali, Hindi, and Spanish, consistently outperforming mBERT and language-specific baselines. Chatterjee et al. \cite{chatterjee2019semeval} selected four emotion classes: happy, sad, angry, and others for the MLE classification. They used a total of 30,160 manually annotated dialogues as a training dataset and tested their models on two different test datasets: Test-1, containing 2,755 and Test-2, containing 5,509 dialogues.  
Many studies have been devoted to developing the highly valuable emotion and sentiment analysis dataset and advanced data-driven machine learning methods for emotion detection. Demszky et al. \cite{demszky2020goemotions} introduced a large-scale manually annotated dataset named \textit{GoEmotions}, which consists of more than 58k comments from various Reddit posts. They labelled 27 emotions (like anger, fear, grief, sadness, love, etc.) categories or neutral, showing that BERT is the strongest baseline. Chakravarthi et al. \cite{chakravarthi2022dravidiancodemix} developed a multilingual dataset named \textit{DravidianCodeMix} for sentiment and offensive language detection in code-mix text on social media. They manually annotated over 60,000 YouTube comment datasets containing over 44,000 in Tamil-English, 70,00 in Kannada-English, and 20,000 in Malayalam-English. Ameer et al. \cite{ameer2022multi} explored code-mixed text in posts, comments, and tweets. A total of 11,914 English and Roman Urdu SMS messages were manually annotated for 12 different emotions, including anger, joy, love, and sadness. Different DL models, such as convolutional neural networks (CNN), recurrent neural networks (RNN), long-short term memory (LSTM), gated recurrent units (GRU) and transfer learning methods (BERT and XLNet) were employed.  

The ability to detect emotion from text is not only significant for sentiment and MLE classification but also useful in the application of mental health. De Choudhury et al. \cite{de2013predicting} have researched how SM posts can reveal the mental health status of the individual. Using Twitter as a platform, they analysed various patterns and language behaviours that could indicate depression or other mental health problems. Their study showed that online activity of an individual on social networks can be used as a valuable tool in the early detection of mental health issues and understanding mental health trends over time. 

For the Nepali language, there have been efforts put forward to develop NLP resources on sentiment analysis \cite{sitaula2021deep, sitoula2025sentiment}, and hate speech detection \cite{thapa2023nehate}. For instance, Rauniyar et al. \cite{rauniyar2023multi} addressed the lack and created a manually annotated Nepal Anti-Establishment Discourse Tweets (NAET) dataset comprising 4,445 multi-aspect tweets in Nepali. In their work, binary classification labels such as relevance, satire detection, hope-speech detection, hate speech detection, and ternary labels, including sentiment and target of hate speech detection, were introduced. Similarly, Thapa et al. \cite{thapa2023nehate} introduced a dataset \textit{NeHate} dedicated to hate speech detection, with approximately 13.5k tweets. 

The summary of the existing datasets on emotion classification, sentiment analysis, and mental health NLP, along with their size, tasks and notable findings, is reported in Table \ref{tab:related_work}. To the best of our knowledge, no prior work has specifically explored MLE classification in Nepali SM text. This is a significant gap as it hinders the development of data-driven models that can effectively capture the complex emotional nuances, particularly in multilingual and code-mixed discourse. 
\begin{table*}[h!]
\centering
\renewcommand{\arraystretch}{2.2}
\caption{Summary of prior datasets and approaches in MLE, SC, and mental health NLP. Our work (last row) uniquely addresses low-resource Nepali Reddit with code-mixed text.}
\begin{tabular}{|c|c|p{1.5cm}|p{3.2cm}|p{3cm}|p{3.5cm}|}
\hline
\textbf{Ref} & \textbf{Year} & \textbf{Size} & \textbf{Tasks} & \textbf{Language / Source} & \textbf{Remarks} \\
\hline
\cite{demszky2020goemotions} & 2020 & $\sim$58k & 27 emotions + Neutral (multi-label) & English / Reddit & BERT fine-tuning; stronger than RNN baselines. \\
\hline
\cite{chatterjee2019semeval} & 2019 & $\sim$38k & Happy / Sad / Angry / Other & English / Dialogues & GRU/attention, later BERT; benchmark dataset. \\
\hline
\cite{sitaula2021deep} & 2021 & $\sim$33k & Sentiment (3-class) & Nepali / Twitter & CNN with multiple embeddings; best performance. \\
\hline
\cite{rauniyar2023multi} & 2023 & $\sim$4.4k & Sentiment, satire, hope-speech, hate & Nepali / Twitter & Multi-aspect annotation; ML/DL baselines compared. \\
\hline
\cite{thapa2023nehate} & 2022/23 & $\sim$13.5k & Hate / Non-hate (+ target labels) & Nepali / Twitter & Annotated dataset; evaluated with Nepali PLMs. \\
\hline
\cite{ameer2022multi} & 2022 & $\sim$11.9k & Emotions (12-class) & Code-mixed / SMS & CNN, BiLSTM, BERT; transformers best for code-mix. \\
\hline
\textbf{Ours} & 2025 & $\sim$4.4k & MLE (Fear, Anger, Sadness, Joy, Depression) + Sentiment (3-class) & Code-mixed (English–Roman Nepali–Devanagari) / Reddit & First Nepali Reddit dataset; temporal analysis; topic modelling;  multilingual transformers outperform ML/DL baselines. \\
\hline
\end{tabular}
\label{tab:related_work}
\end{table*}

\section{Data collection and annotation}
\label{dataset}
\begin{figure}[h!]
    \centering
    \includegraphics[width=\linewidth]{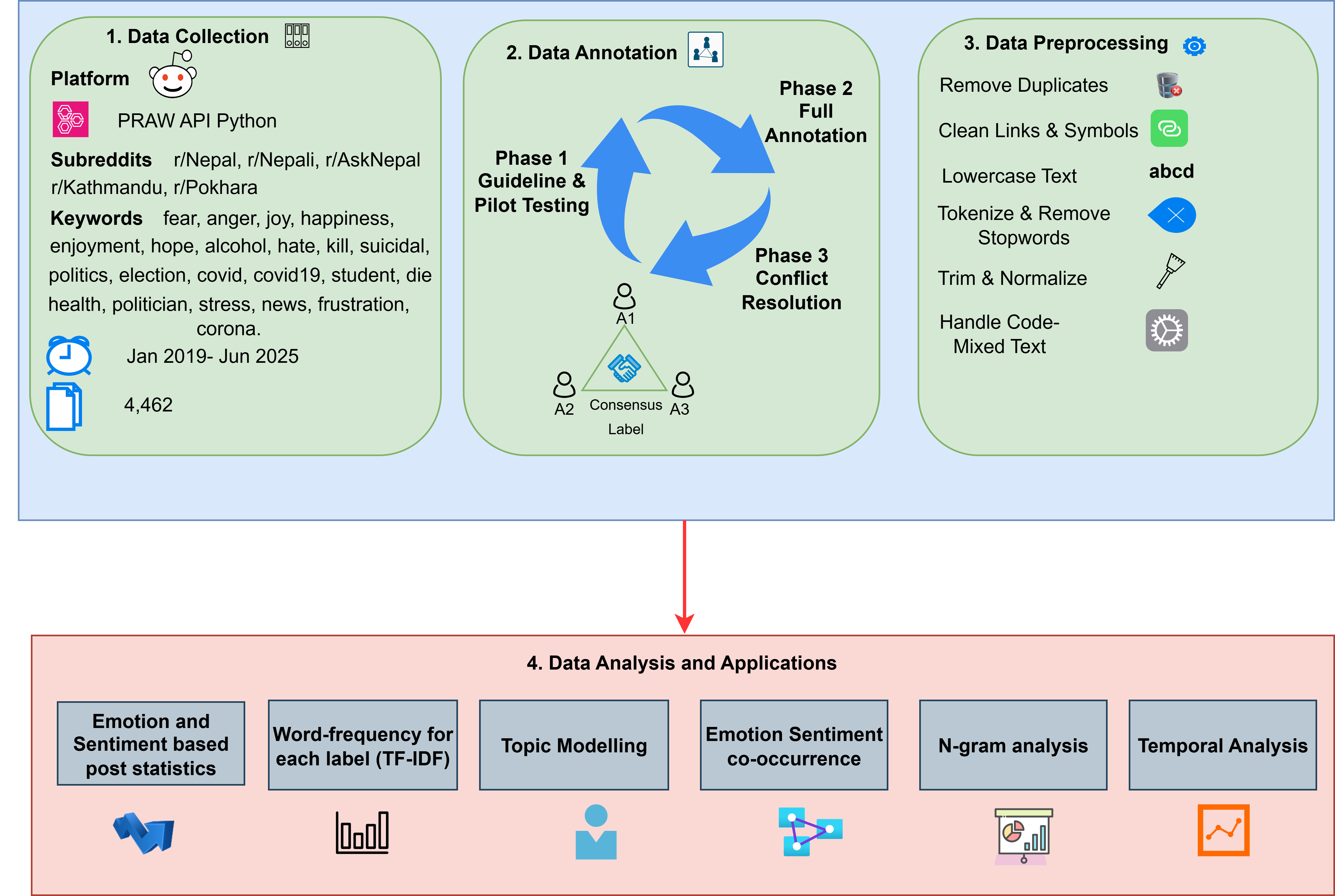} 
    \caption{Architecture of our methodology, including data collection, annotation, preprocessing, and analysis}
    \label{fig:architecture}
\end{figure}

Here, we outline the procedure followed to create a NepEMO dataset, including data collection, annotation, and task definitions. Fig. \ref{fig:architecture} illustrates the methodology for collecting, annotating, and analysing multilingual NepEMO posts, gathered using keywords for various subreddits.

\subsection{Data collection}
To ensure the anonymity of the user and collect reliable data, Reddit was selected as the primary source for data crawling. The PRAW API \footnote{\url{https://praw.readthedocs.io/}, accessed on September 18, 2025} for Python is used to crawl the Reddit post. We utilised various subreddits, namely \texttt{ r/Nepal}, \texttt{r/Nepali}, \texttt{r/Kathmandu}, \texttt{r/Pokhara} and \texttt{r/AskNepal} and specific keywords related to mental health, specifically \texttt{stress, fear, anger, joy, happiness, enjoyment, hope, alcohol, hate, kill, suicidal, politics, election, covid, covid19, student, health, die,  politician, news, frustration, and corona}.

With these keywords and subreddits, we approximately acquired 5154 posts, spanning from \textbf{January 1 2019} to \textbf{June 5, 2025} (nearly six and a half years duration). This time span is critical for mental health discussion on social media as it includes the time of the global pandemic (COVID-19), the 2022 parliamentary elections, local elections, the current trend of students shifting abroad for their higher studies, and various protests. After filtering the duplicate samples, we end up with a dataset of 4,462 Reddit posts. Therefore, the dataset contains Reddit posts that signal the people's mental health and well-being, and their conversations about these issues in the online community. 

\subsection{Annotation process}
The data annotation process involves assigning a specific label to each sample. The labelled dataset is crucial for textual analysis and model development. The NepEMO dataset is annotated through a rigorous process in which three native speakers of Nepali, each with at least a master's degree and English language as the medium of instruction, were employed. The annotators were supplied with clear and concise data annotation guidelines, which helped ensure the reliability and validity of the dataset. The details of the data annotation process are as follows.


\subsubsection{Three-phase annotation}
The annotation process was conducted in three phases: a) Annotation guideline understanding and pilot testing, b) Full annotation, and c) Conflict resolution. Initially, instructions and guidelines on how to annotate and provide a label for each text were developed. Then, the pilot run was conducted to ensure that everyone had understood the annotation rules and guidelines. This also provides a chance to revise the guidelines if any inconsistencies are found during this phase. Second, the full annotation was performed for all samples, and finally, the conflict resolution phase was conducted to filter out inconsistent labelling and re-assign the label.

\textbf{First phase:}
In this phase, all three annotators collectively discussed and labelled the first 100 posts. Each post was carefully read, and annotators engaged in discussion based on the established guidelines (Section \ref{guidlines}). Here, the goal was to label Reddit posts for five multi-label emotions and three sentiment classes. Any uncertainties regarding annotation or labelling were addressed by revising the guidelines. This phase ensured that all annotators fully understood the process and were aligned on the annotation standards. 

\textbf{Second phase:}
In this phase, all annotators appropriately labelled the dataset based on the discussion from the first phase and the finalised guidelines. Each annotator also assigned a trust score to each post, indicating their confidence in the accuracy of the label. Trust scores were given as follows: 1.0 for full confidence, between 0.9 and 0.99 for moderate confidence, and below 0.90 for uncertainty. This phase concluded the primary annotation process.

\textbf{Third Phase:}
This phase focused on the conflict resolution phase. The posts with trust scores of less than 0.90 were again discussed by all three annotators collectively and re-labelled. The overall aim was to ensure that all annotations were unambiguous. This phase was essential for resolving disagreements and clarifying any uncertainties. A total of 670 posts had received trust scores below 0.90, which were corrected and re-annotated during this phase.  

\subsubsection{Annotation guidelines}
\label{guidlines}
A comprehensive guide was developed and provided to the annotation team to ensure reliability and uniformity in the annotation process. These guidelines were then used to manually annotate the Reddit post in English and code-mixed (English, Roman Nepali, and Devanagari) for MLE classification. Because the task involved multi-label emotion classification, all annotators were asked to assign multiple labels or emotions, in our case. We followed a similar process used by \cite{demszky2020goemotions, ameer2022multi,chatterjee2019semeval, naseem2022early,rauniyar2023multi} for the annotation of emotion and sentiment labels. Following the guidelines, discussion, and revision made in the first phase, the three annotators assigned one or more emotions to the datasets of the following six categories. We used a Google sheet to perform the annotation process, and it was completed in about two months.

\textbf{ i) Annotation guidelines for emotion categories}
\begin{itemize}
   \item [a)] \textbf{Fear:} The emotion fear is known when a person is scared, worried, or anxious. It appears when an individual is threatened or uncertain about a situation in which they are, and it often leads to negative outcomes. In our annotation, when a post has these instances, they were labelled as Yes and No if they are not present. 
   \item [b)]\textbf{Anger:} It is an intense emotional response when an individual is frustrated, hostile, irritated, or feels the injustice faced. If not properly controlled and handled, it can lead to major harmful actions. In our annotation, when a post has these instances, they were labelled as Yes and No if they are not present. 
   \item [c)] \textbf{Sadness:} When an individual experiences disappointment, loss, emptiness, or lack of motivation, they tend to experience this emotion. It is a negative state when a person processes grief in various experiences they face. In our annotation, when a post has these instances, they were labelled as Yes and No if they are not present. 
   \item [d)]\textbf{Joy:} This is a positive emotion when an individual feels happy and satisfied. This is often associated with success, hope, and fulfillment and leads to optimism and improved overall well-being. In our annotation, when a post has these instances, they were labelled as Yes and No if they are not present. 
   \item [e)]\textbf{Depression:} This goes beyond sadness or fear emotion and often leads to suicidal thoughts, lack of interest in daily activities, and lack of worth. Unlike short-term emotional distress, depression can affect relationships, quality of life, and significantly affect the mental health of the individual. In our annotation, when a post has these instances, they were labelled as Yes and No if they are not present. \\
\end{itemize}
\vspace{5mm}

\textbf{ii)Annotation guidelines for sentiment categories}

Sentiment analysis (SA) is significant in understanding the overall emotion conveyed by the post. Annotators are instructed to meticulously assess each post's content and label whether the overall sentiment is \textit{positive}, \textit{negative}, or \textit{neutral}. The focus should be on general sentiment rather than individual emotion. For example, a person may initially express sadness or anger that can lead to depressive thoughts later, and then later share, with the proper guidance of a mentor, parents, or friends, overcame these negative thoughts and achieved joy and happiness later in their life. Therefore, the overall sentiment should be provided based on the concluding emotional state. 

\begin{itemize}
    \item [a)] \textbf{Identifying positive sentiment:} The overall sentiment will be positive when a post expresses hope, happiness, satisfaction, or optimism. This can be expressed in terms of success, encouragement, or recovery from difficult emotions.

    \item [b)] \textbf{Identifying negative sentiment:} When a post is expressed with anger, sadness, hopelessness, dissatisfaction, criticism, complaints, depression, fear, anxiety, and frustration, it is termed negative. 

    \item [c)]\textbf{Identifying neutral sentiment:} A post that does not contain any strong emotion or objective is termed neutral. It may also contain general information, asking for guidance or announcements, and it does not lean strongly towards positive or negative. Also, sarcastic and mixed-feeling posts are annotated as neutral sentiment. 
\end{itemize}

Table \ref{tab:emo/sent_examples} lists a few examples of annotated Reddit posts, each containing one or more emotions and the corresponding sentiment labels.

\begin{table}[!h]
\centering
\renewcommand{\arraystretch}{2.5}
\caption{Examples of original texts with corresponding multi-label emotions separated by commas and Sentiment labels (positive, negative or neutral).}
\label{tab:emo/sent_examples}
\begin{tabular}{|c|p{7cm}|p{4.0cm}|}
\hline
\textbf{S.N.} & \textbf{Reddit post (self-text)} & \textbf{Emotion / Sentiment } \\ \hline
1 & Please mind my English.
Its 5:30 . I haven’t slept whole night worried 
I am shy and introvert guy with low self esteem.
I passed my +2 in 2017 and I went to Australia. I stayed there for 2.5 yrs . During covid I struggled with depression and returned back home with debt. I joined Bsc physics and part time teaching earning 15k (9-4) . But I am looking to start anything other than this life. Please Any suggestions will be greatly appreciated. & Fear, Sadness, Depression / Negative \\ \hline
 2 & I am +2 student and due to covid online classes are being conducted. Today in my home town it was really cold therefore I stay whole day in my room and went to gym at 5PM, I was feeling depressed idk why but i just wanted to do nothing like i just cant express how I was feeling... But after  my workout session I am feeling quite good. So, was the feeling occuring coz of lack of physical activity or what? & Sadness, Joy, Depression / Neutral \\ \hline

3 & Depression is REAL! Made a video on depression, its causes and signs. I have also talked about the history of mental health in Nepal. Would love some suggestions. Thank you. & Joy / Positive \\ \hline
\end{tabular}
\end{table}

\subsubsection{ Inter-annotator agreement and statistics}
It is very important to validate the reliability of annotation between multiple annotators. To evaluate agreement between annotators, we used Fleiss'($\kappa$) \cite{fleiss1971measuring} as defined in Eq. \ref{eq:pj_pi}.

For items $N$ annotated by $3$ annotators in categories $k$, let $n_{ij}$ denote the number of annotators who assigned the item $i$ to category $j$

{\small
\begin{equation}
p_j = \frac{1}{3N} \sum_{i=1}^{N} n_{ij}, \quad
P_i = \frac{1}{3 \cdot 2} \left( \sum_{j=1}^{k} n_{ij}^2 - 3 \right)
\label{eq:pj_pi}
\end{equation}
}
where $p_j$ is the proportion of all assignments in category $j$ and $P_i$ is the agreement for item $i$.  

The mean observed agreement and expected agreement by chance are defined as in Eq.\ref{eq:mean_agreement}.

{\small
\begin{equation}
\bar{P} = \frac{1}{N} \sum_{i=1}^{N} P_i, \quad
\bar{P_e} = \sum_{j=1}^{k} p_j^2
\label{eq:mean_agreement}
\end{equation}
}
Finally, Fleiss' Kappa($\kappa$) is computed as in Eq.\ref{eq:fleiss_kappa}.
{\small
\begin{equation}
\kappa = \frac{\bar{P} - \bar{P_e}}{1 - \bar{P_e}}
\label{eq:fleiss_kappa}
\end{equation}
}

A value of $\kappa = 1$ indicates perfect agreement, $\kappa = 0.81 - 0.99$ indicates almost perfect agreement, $\kappa = 0.61 - 0.80$ indicates good agreement, $\kappa = 0$ indicates agreement equivalent to chance, and $\kappa < 0$ indicates worse agreement \cite{fleiss1971measuring} .

Table \ref{tab:fleiss_kappa_values} shows the Fleiss' Kappa ($\kappa$) score obtained for all three annotators on emotion and sentiment labels. The results in Table \ref{tab:fleiss_kappa_values} indicate that we have achieved almost perfect agreement in annotation. The reason is that we used multiple phases of annotation which helped to annotate the posts rigorously and resolve conflicts that arise during the annotation.

\begin{table}[!h]
\centering
\renewcommand{\arraystretch}{1.5}
\caption{Fleiss' Kappa ($\kappa$) Values for Emotions and Sentiment. Boldface indicates the highest score.}
\label{tab:fleiss_kappa_values}
\begin{tabular}{|l|c|c|c|c|c|c|}
\hline
\textbf{Emotion / Sentiment} & \textbf{Fear} & \textbf{Anger} & \textbf{Sadness} & \textbf{Joy} & \textbf{Depression} & \textbf{Sentiment} \\ \hline
\textbf{Fleiss' Kappa ($\kappa$)} & 0.8820 & \textbf{0.9431} & 0.9050 & 0.9238 & 0.8938 & 0.8674 \\ \hline
\end{tabular}
\end{table}

\begin{table}[h!]
\centering
\renewcommand{\arraystretch}{2.5} 
\setlength{\tabcolsep}{11pt} 
\caption{Statistics of the NepEMO dataset for emotions and sentiment classification tasks}
\label{tab:label_distribution}
\begin{tabular}{|l|c|c|c|}
\hline
\textbf{Emotion \& sentiment} & \textbf{Label} & \textbf{Posts} & \textbf{Avg.Words}\\ \hline
Joy & Joy & 1195 (26.74\%) & 46.553 \\ 
    & No Joy & 3267 (73.26\%) & 57.003  \\ \hline
Anger & Anger & 1369 (30.72\%) & 82.290\\ 
      & No Anger & 3093 (69.28\%) & 41.773 \\ \hline
Sadness & Sadness & 1744 (39.09\%) & 85.843 \\ 
        & No Sadness & 2718 (60.91\%) & 33.903 \\ \hline
Fear & Fear & 1625 (36.39\%) & 77.163 \\ 
     & No Fear & 2837 (63.61\%) & 41.054\\ \hline
Depression & Depression & 949 (21.20\%) & 102.457 \\ 
           & No Depression & 3513 (78.80\%) & 41.169 \\ \hline
Sentiment & Positive & 1040 (24.46\%) & 40.827 \\ 
          & Neutral & 1261 (29.66\%) & 35.302 \\ 
          & Negative & 2161 (45.29\%) & 71.672 \\ \hline
\end{tabular}
\end{table}

\subsection{Data analysis}
\label{exploratory_analysis}
\subsubsection{Dataset statistics}
\label{dataset_statistics}
The statistics of the NepEMO dataset are reported in Table \ref{tab:label_distribution}. For the MLE task, \textit{Sadness} has the highest occurrence with 1744 posts (39.09 \%), followed by \textit{Fear} with  1625 (36.39\%) posts. The least frequent emotion is Depression, with only 949 posts (21.20\%) labelled as \textit{Depression}. For the SC task, the dataset includes 1,040 (24.46\%) posts as positive, 2,161 (45.29\%) posts as negative, and 1,261 (29.66\%) posts as neutral sentiments.
As shown in Table \ref{tab:label_distribution}, posts that have the emotion label as \textit{Anger}, \textit{Sadness}, \textit{Fear}, \textit{Depression}, have an average word length of almost twice or more than that of posts that are labelled as not having these emotions \cite{aleksandric2023sadness, liu2022head}. This might imply that people who feel low, disappointed, worried, and irritated tend to write and express longer texts. 


\begin{table}[!h]
\centering
\caption{Title and selftext present in our NepEMO dataset}
\label{tab:title_selftext}
\begin{tabular}{|c|p{2cm}|p{7cm}|c|}
\toprule
\textbf{S.No} & \textbf{Title} & \textbf{Self-text (post)}&\textbf{Language} \\ 
\midrule
1 & Health help & So there's no vacant room with required monitor in Patan hospital, they are referring to Teaching hospital for the room. We have to vacat the 24-hr emergency room tonight. So what should we do. Will there be room available in Teaching hospital. Will there be a toll to the patient while switching hospital. What should we do, any helpful suggestions are welcomed.&English \\ 
\hline
2 & First Breakup & Had my first breakup \textit{3 months ko relationships theo tei pani long distance tara and long distance re}(roughly translated as ''it was a long distance relationship of three months'')  still i’m having hard times to accept she is gone.I’m struggling with breakup anxiety need some help \textit{saaja ko time maa ali garo hunxa}(roughly translated as 'it is  a bit hard in evening time').&Code-mixed  
\\
\bottomrule

\end{tabular}
\end{table}

\subsubsection{Language distribution}
\label{language_distribution}
\begin{figure}[h!]
    \centering
    \includegraphics[width=0.5\textwidth]{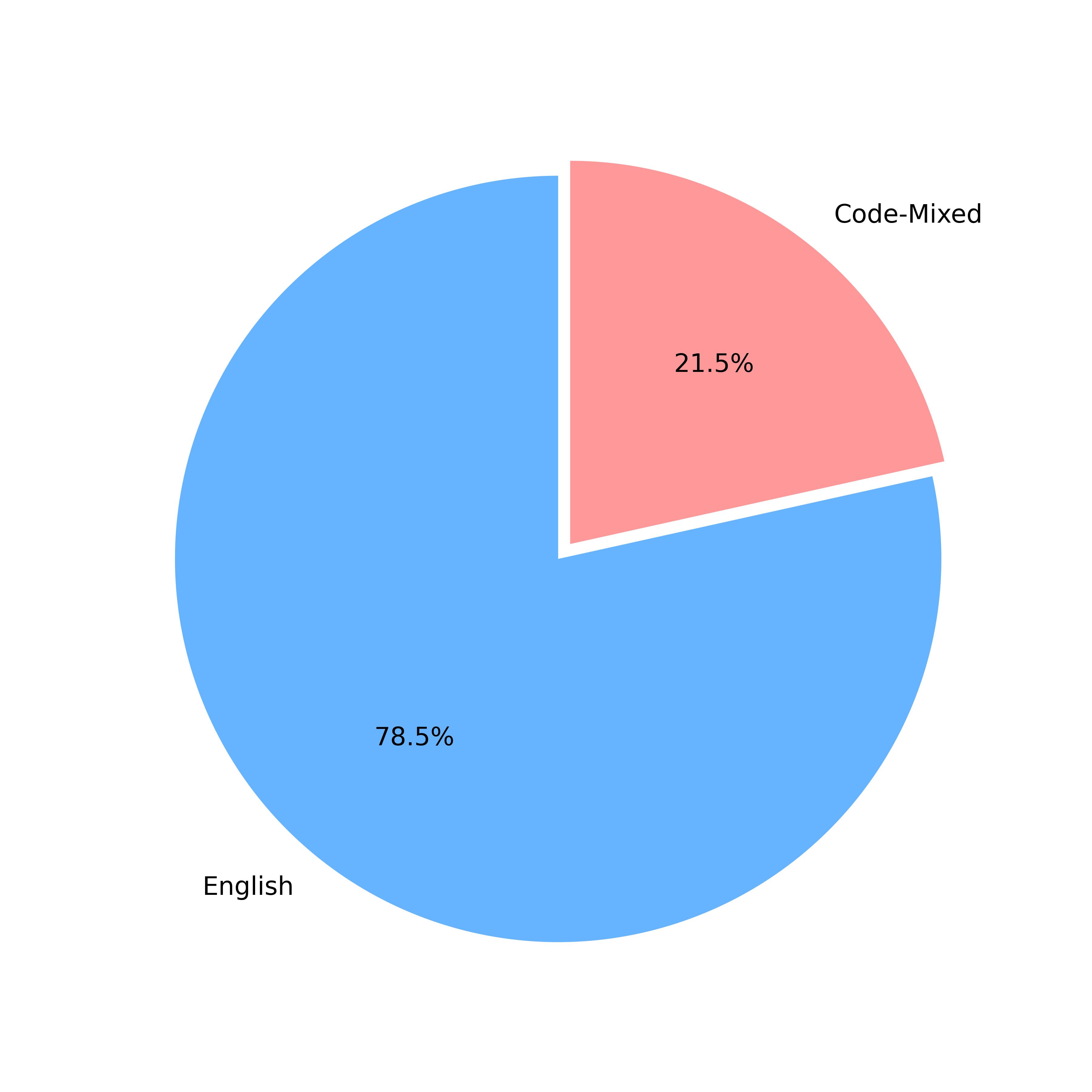} 
    \caption{Language distribution of the dataset: English vs Code-Mixed posts.}
    \label{fig:language_distribution}
\end{figure}

Among the total 4,462 Reddit posts in NepEMO dataset, 3,501 (78.5\%) are entirely in English and 961 (21.5\%) are code-mixed languages. Here, code-mixed languages include the post written in English-Nepali, Roman Nepali mixed with English, and all three mixed. The statistics are shown in Fig. \ref{fig:language_distribution}. Example posts written in English and code-mixed language from the NepEMO dataset, along with titles, are shown in Table \ref{tab:title_selftext}.

\subsubsection{Word frequency distribution}
\label{word_freq_distribution}
Here, we use the Term Frequency-Inverse Document Frequency (TF-IDF) to extract the relevant words according to the weight. This is performed on a preprocessed dataset and can be computed as follows in Eq.\ref{eq:tfidf_stopwords}: 

{\small
\begin{equation}
\text{TF-IDF}_{\text{filtered}}(w,d,D,S) =
\begin{cases}
\text{TF}(w,d) \times \log \frac{|D|}{1 + |\{d'' \in D : w \in d''\}|}, & \text{if } w \notin S \\
0, & \text{if } w \in S
\end{cases}
\label{eq:tfidf_stopwords}
\end{equation}
}
The average word across all documents using TF-IDF can be shown as in Eq.\ref{eq:avg_tfidf}
{\small
\begin{equation}
\overline{\text{TF-IDF}}(w,D,S) = \frac{1}{|D|} \sum_{d \in D} \text{TF-IDF}_{\text{filtered}}(w,d,D,S)
\label{eq:avg_tfidf}
\end{equation}
}
where $w$, $d$, $D$ and $S$ represent word (term), individual document, and set of all documents and set of stopwords (English, Roman Nepali, Devanagari Nepali) respectively.
Table \ref{tab:tf-idf_emotions} shows the information for the categories: No-Fear, Fear, No-Anger, Anger, No-Sadness, Sadness, No-Joy, Joy, No-Depression and Depression and finally for the Positive, Negative, and Neutral sentiments in Table \ref{tab:tf-idf_sentiment}. As seen in Table \ref{tab:tf-idf_emotions} and Table \ref{tab:tf-idf_sentiment}, the terms such as ``nepal'', ``people'' and ``like'' are significant in all the tasks. Whereas, for negative emotions such as ( Fear, Anger, Sadness, No Joy, Depression and Negative Sentiment), terms such as ``like'', ``even'', ``get'', and ``one'' are repeated in all negative emotion task and some terms related to emotional strain such as ``feel'' and ``time'' are repeated in most. Similarly, for positive emotions such as (No-Fear, No-Anger, No-Sadness, Joy, No-depression, and Positive sentiment), terms such as ``happy'', ``hope'', ``news'' and ``like'' are repeated frequently. And there is frequent repetition of ``enjoy'', ``good'', ``nepali'', and ``day'' in Joy and Positive sentiment posts.

\begin{table}[!h]
\centering
\caption{Top 10 most frequent terms for Fear, Anger, Sadness, Joy, and Depression labels}
\begin{tabular}{|l l||l l||l l||l l||l l|}
\hline\hline
\multicolumn{2}{|c||}{\textbf{No Fear}} & \multicolumn{2}{c||}{\textbf{Fear}} & \multicolumn{2}{c||}{\textbf{No Anger}} & \multicolumn{2}{c||}{\textbf{Anger}} & \multicolumn{2}{c|}{\textbf{No Sadness}} \\
\hline
\textbf{Term} & \textbf{TF-IDF} & \textbf{Term} & \textbf{TF-IDF} & \textbf{Term} & \textbf{TF-IDF} & \textbf{Term} & \textbf{TF-IDF} & \textbf{Term} & \textbf{TF-IDF} \\
\hline\hline
nepal & 0.041 & nepal & 0.045 & nepal & 0.044 & nepal & 0.042 & nepal & 0.044 \\
news & 0.026 & like & 0.034 & like & 0.025 & people & 0.038 & news & 0.026 \\
happy & 0.024 & people & 0.030 & covid & 0.023 & like & 0.033 & covid & 0.023 \\
people & 0.024 & covid & 0.026 & happy & 0.022 & even & 0.024 & happy & 0.023 \\
like & 0.024 & know & 0.025 & news & 0.022 & nepali & 0.023 & like & 0.023 \\
hope & 0.021 & even & 0.025 & hope & 0.021 & news & 0.022 & hope & 0.021 \\
nepali & 0.020 & get & 0.023 & people & 0.020 & get & 0.020 & people & 0.020 \\
get & 0.017 & time & 0.023 & corona & 0.020 & know & 0.020 & corona & 0.020 \\
one & 0.017 & one & 0.019 & know & 0.019 & one & 0.020 & nepali & 0.018 \\
would & 0.016 & corona & 0.019 & get & 0.019 & country & 0.020 & get & 0.017 \\
\hline\hline
\multicolumn{2}{|c||}{\textbf{Sadness}} & \multicolumn{2}{c||}{\textbf{No Joy}} & \multicolumn{2}{c||}{\textbf{Joy}} & \multicolumn{2}{c||}{\textbf{No Depression}} & \multicolumn{2}{c|}{\textbf{Depression}} \\
\hline
\textbf{Term} & \textbf{TF-IDF} & \textbf{Term} & \textbf{TF-IDF} & \textbf{Term} & \textbf{TF-IDF} & \textbf{Term} & \textbf{TF-IDF} & \textbf{Term} & \textbf{TF-IDF} \\
\hline\hline
nepal & 0.040 & nepal & 0.043 & happy & 0.046 & nepal & 0.044 & like & 0.040 \\
people & 0.034 & people & 0.029 & nepal & 0.043 & news & 0.025 & nepal & 0.038 \\
like & 0.034 & like & 0.028 & hope & 0.033 & people & 0.024 & people & 0.033 \\
even & 0.026 & covid & 0.022 & enjoy & 0.029 & like & 0.024 & know & 0.030 \\
know & 0.025 & get & 0.022 & like & 0.026 & covid & 0.022 & even & 0.029 \\
time & 0.023 & news & 0.021 & news & 0.024 & happy & 0.022 & life & 0.028 \\
life & 0.023 & know & 0.021 & nepali & 0.022 & hope & 0.020 & feel & 0.028 \\
get & 0.023 & even & 0.019 & guys & 0.021 & nepali & 0.019 & time & 0.026 \\
feel & 0.022 & one & 0.018 & good & 0.021 & corona & 0.019 & get & 0.026 \\
one & 0.021 & time & 0.018 & day & 0.021 & get & 0.017 & one & 0.023 \\
\hline\hline
\end{tabular}
\label{tab:tf-idf_emotions}
\end{table}

\begin{table}[!h]

\centering
\caption{Top 10 most frequent terms for sentiment labels: Positive, Negative, and Neutral}
\begin{tabular}{|l l||l l||l l|}
\hline\hline
\multicolumn{2}{|c||}{\textbf{Neutral}} & \multicolumn{2}{c||}{\textbf{Positive}} & \multicolumn{2}{c|}{\textbf{Negative}} \\
\hline
\textbf{Term} & \textbf{TF-IDF} & \textbf{Term} & \textbf{TF-IDF} & \textbf{Term} & \textbf{TF-IDF} \\
\hline\hline
nepal & 0.045 & happy & 0.048 & nepal & 0.040 \\
covid & 0.025 & nepal & 0.044 & like & 0.033 \\
news & 0.023 & hope & 0.035 & people & 0.033 \\
get & 0.022 & enjoy & 0.030 & even & 0.024 \\
corona & 0.021 & news & 0.025 & know & 0.022 \\
people & 0.021 & like & 0.025 & get & 0.021 \\
like & 0.020 & nepali & 0.022 & time & 0.020 \\
know & 0.020 & guys & 0.022 & news & 0.020 \\
anyone & 0.018 & day & 0.022 & one & 0.019 \\
health & 0.017 & good & 0.022 & nepali & 0.019 \\
\hline\hline
\end{tabular}
\label{tab:tf-idf_sentiment}
\end{table}

\subsubsection{Unsupervised topic modelling and theme extraction}

Topic modelling is an unsupervised learning method. It helps uncover hidden themes from a large collection of textual data. Topic modelling helps us generate topics with similar themes and share words that convey the same meaning. Thus, topic modelling is based on three main ideas: \textbf{words}, \textbf{documents}, and \textbf{corpora}. A word is the smallest unit in text, a document is a collection of words, and a corpus is a collection of multiple documents. Among various methods, Latent Dirichlet Allocation (LDA) is one of the most popular methods that was first introduced by Blei et al. \cite{blei2003latent}. It assumes that documents contain a mixture of topics and each topic is represented by a distribution of words. Therefore, it is very useful for SM topic extraction, where posts contain multiple overlapping themes, and it is applied to identify topics related to socio-political and emotional discussion \cite{tong2016text}.  The process can be expressed as \cite{blei2003latent} in Eq.\ref{eq:lda}: 

{\small
\begin{equation}
\begin{aligned}
\theta_d &\sim \text{Dirichlet}(\alpha), \quad d = 1, \dots, M\\
\phi_k &\sim \text{Dirichlet}(\beta), \quad k = 1, \dots, K\\
z_{dn} &\sim \text{Categorical}(\theta_d), \quad n = 1, \dots, N_d\\
w_{dn} &\sim \text{Categorical}(\phi_{z_{dn}})
\end{aligned}
\label{eq:lda}
\end{equation}
}
{\small
where $M$, $N_d$, and $K$, represent the number of documents, words, and topics, respectively. And, $\theta_d$, and $\phi_k$, denote the topic and word distribution for document $d$, respectively. And, $z_{dn}$  and $w_{dn}$ denote the topic assignment for the $n$-th word in document $d$, and the $n$-th word in document $d$, respectively.



Fig. \ref{fig:coherence_score} shows the number of optimal topics compared to the C\_v coherence score \cite{roder2015exploring, hasan2020normalized}. Here, we can see that for the number\_of\_topics = 2, the coherence score is computed as 0.465, which gradually increased and reached around 0.505 for 3 topics. However, it again dropped slightly by around the score of 0.017 and reached 0.488 before rising sharply to 0.514 for five topics. In general, the coherence score fluctuates in the range of topics, as shown in Fig. \ref{fig:coherence_score} up to 14 topics. 

\begin{figure}[h!]
    \centering
    \includegraphics[width=0.5\textwidth]{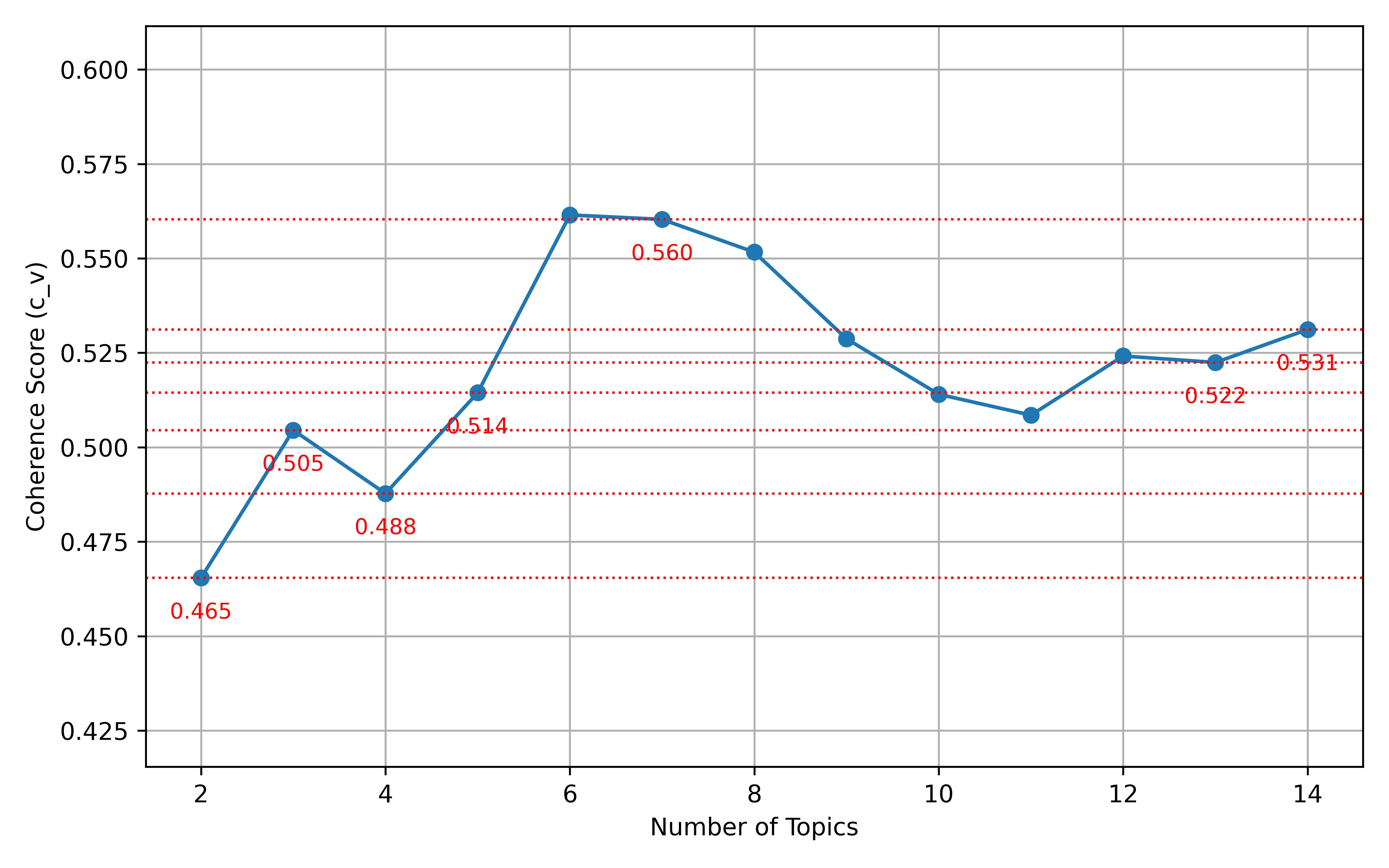} 
    \caption{Number of topics compared to coherence score}
    \label{fig:coherence_score}
\end{figure}

Thus, we evaluated topic modelling based on the coherence score. The highest coherence score is achieved when the number of topics is 6. However, as shown in Fig. \ref{fig:lda_topic_6}, 6 topics have been chosen. However, the topics overlap with each other. Similarly, for the number of topics as 5 as shown in Figure \ref{fig:lda_topic_5}, a similar cluster overlapping problem is repeated. Therefore, both have been discarded. Finally, after a few more tests, we found that using the number of topics as 3 generated a good partition in each quartile and did not overlap. Figure \ref{fig:lda_topic_3} shows that 3 is the number of topics that have been selected having a coherence score of 0.505 \cite{hasan2020normalized}.

\begin{figure}[h!]
    \centering
    \includegraphics[width=\linewidth]{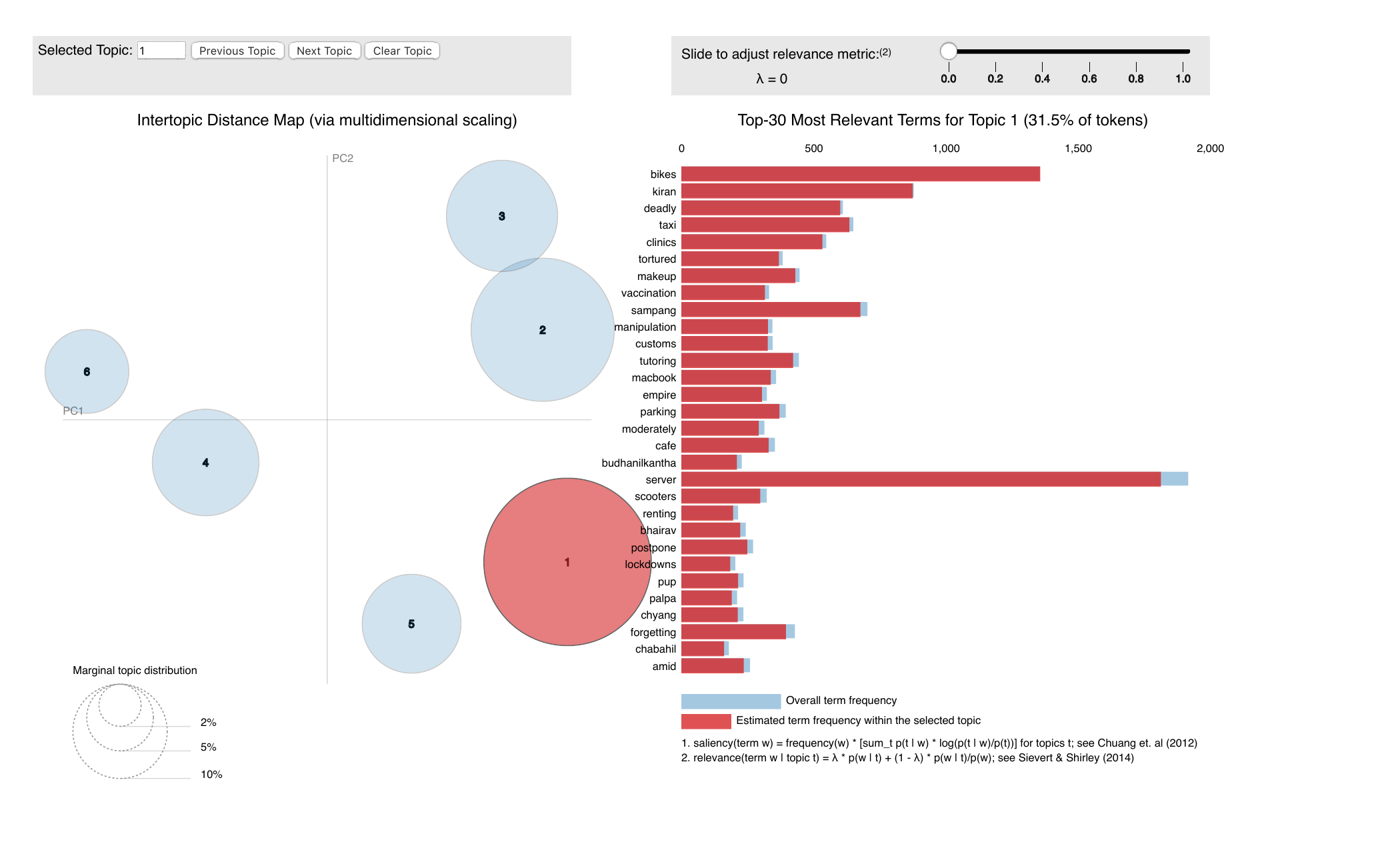} 
    \caption{Rejected topic modelling due to overlapping(no.of topics = 6)}
    \label{fig:lda_topic_6}
\end{figure}

\begin{figure}[h!]
    \centering
    \includegraphics[width=\linewidth]{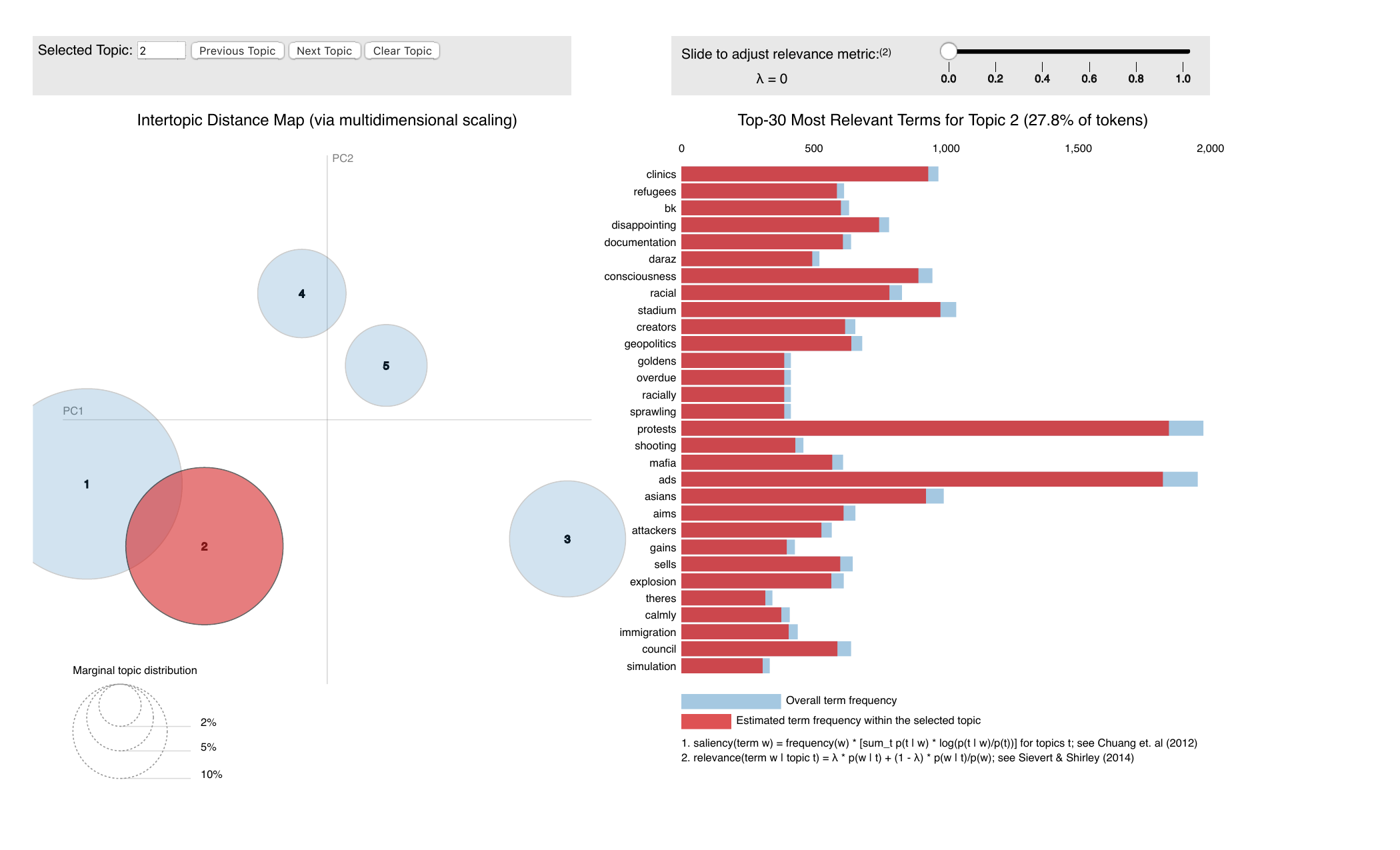} 
    \caption{Rejected topic modelling due to overlapping(no.of topics = 5)}
    \label{fig:lda_topic_5}
\end{figure}

\begin{figure}[h!]
    \centering
    \includegraphics[width=\linewidth]{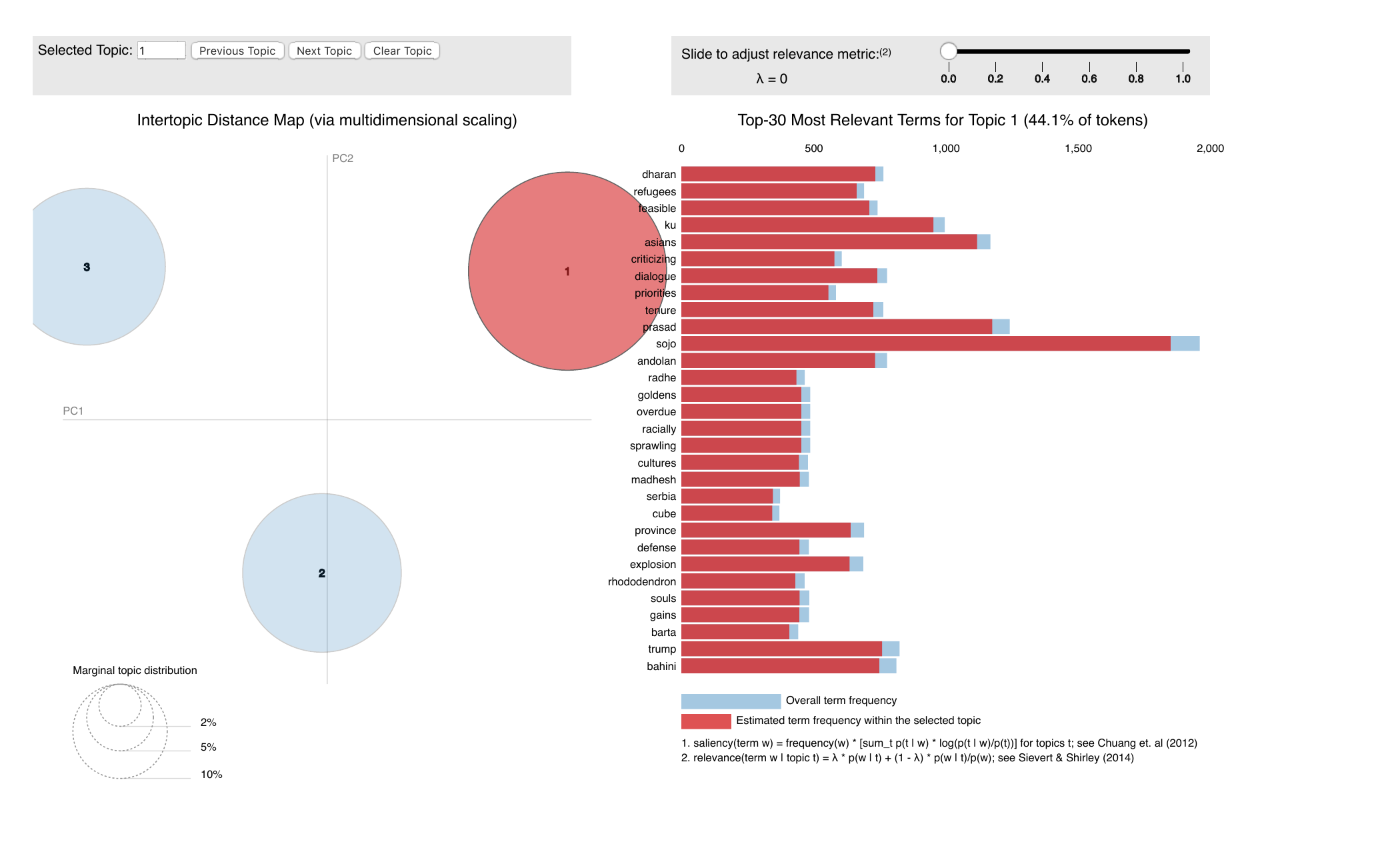} 
    \caption{Selected topic modelling (no.of topics = 3)}
    \label{fig:lda_topic_3}
\end{figure}

Here, we visualised the topic modelling with a $\lambda$ of zero, as shown in Figs. \ref{fig:lda_topic_6}, \ref{fig:lda_topic_5} and \ref{fig:lda_topic_3}. When the value of $\lambda$ is set to zero, it shows the frequency of the estimated term for the selected topic, unlike $\lambda = 1$ or other values of $\lambda$ showing the frequency of the overall term \cite{sievert2014ldavis}. Hence, when the value of $\lambda$ is set to zero, it can be helpful to determine the correct domain for certain topics based on their frequent words.

The relevance of a term $w$ for a topic $t$ at a given $\lambda$ is defined as \cite{sievert2014ldavis} in Eq.\ref{eq:relevance}:

{\small
\begin{equation}
r(w, t \mid \lambda) = \lambda \, p(w \mid t) + (1 - \lambda) \, \frac{p(w \mid t)}{p(w)}
\label{eq:relevance}
\end{equation}
}
where $p(w)$ and $p(w \mid t)$  marginal probability of word $w$ in the corpus and the probability of word $w$ given topic $t$, respectively. And $\lambda \in [0,1]$ controls the trade-off between term frequency and term distinctiveness. 


\begin{table}[!h]
\centering
\caption{Topic modelling results (3 topics) with domain interpretation and most frequent terms.}
\label{tab:topic_modelling}
\begin{tabular}{|p{0.7cm}|p{3cm}|p{6.5cm}|}
\hline
\textbf{Topic} & \textbf{Domain Name} & \textbf{Frequent Terms (with translation)} \\ \hline

1 & Politics, Migration, and Social Movements  & Dharan, refugees, feasible, Asians, criticizing, dialogue, priorities, tenure, Prasad, andolan (movement/protest), racially, cultures, Madhesh, Serbia, province, defense, explosion, Trump, bahini (younger sister) \\ \hline

2 & Tourism, Religion, and Cultural Practices & Buddhism, Kalinchowk, CPN, Qatar, menstruation, vaccines, visibility, butter, woolen, Ranipauwa, Poonhill, hut, journalists, paragliding, Aila (local liquor), retreat, Gautam (Buddha), Budhanilkantha, Buddhist, EBC (Everest Base Camp) \\ \hline

3 & Health, Education, and Technology  & clinics, banking, improvements, SSD, stigma, requirement, hamile (we), dose, Pokharama (in Pokhara), pharmacies, renewal, wine, ACCA, Lenovo, MacBook, graphics, HIV, compulsory, monitor, registered \\ \hline
\end{tabular}
\end{table}

Table \ref{tab:topic_modelling} shows the results of the topic modelling performed on Nepali Reddit posts. Our analysis revealed three distinct topics, each of which represents a major theme discussed by the users.

\begin{itemize}
\item [i)] \textbf{Politics, Migration, and Social movements:} Conversations on local politics, social movements, and migration issues are included in this topic. Words such as \textit{Dharan}, \textit{refugees}, and \textit{andolan (movement/protest)} demonstrate that people actively engage in social and political discussions. References to global events and personalities, as \textit{Trump} and \textit{Serbia}, imply that users connect local problems to global settings. 

\item [ii)]\textbf{Tourism, Religion, and Cultural practices:} The topic focuses on Nepal's tourism industry, religious customs, and traditions. Words such as \textit{Buddhism}, \textit{Buddha}, \textit{Paragliding}, \textit{Kalinchowk},\textit{Poonhill}, \textit{EBC (Everest Base Camp)}, and \textit{Aila (local liquor)} indicate conversations about both tradition, religion and travel. This highlights how people use social networks to spread cultural knowledge and advertise tourism.

\item [iii)]\textbf{Health, Education, and Technology:} The discussion here reflects the conversation related to technology, education, and health services. Terms like \textit{clinics}, \textit{HIV}, \textit{dose}, \textit{improvement}, \textit{MacBook}, \textit{Lenovo}, and \textit{Pokharama (in Pokhara)} express concerns about modern technology, education, and access to healthcare. Similarly, terms like \textit{compulsory} and \textit{stigma} demonstrate conversations on rules and social issues.

\end{itemize}
\subsubsection{Co-occurrence of emotions}
\label{co-occurrence}

In this section, we show how many posts contain one or more labels and how many are related to each other. In Table \ref{tab:emo/sent_examples}, we show that one single post can have multiple emotions, and certain posts can also have anger, fear, joy, and depression. To further clarify this part, we show a network diagram that gives the co-occurrence of emotions as shown in Figure \ref{fig:emotion_cooccurrence}.

\begin{figure}[h!]
    \centering
    \includegraphics[width=\linewidth]{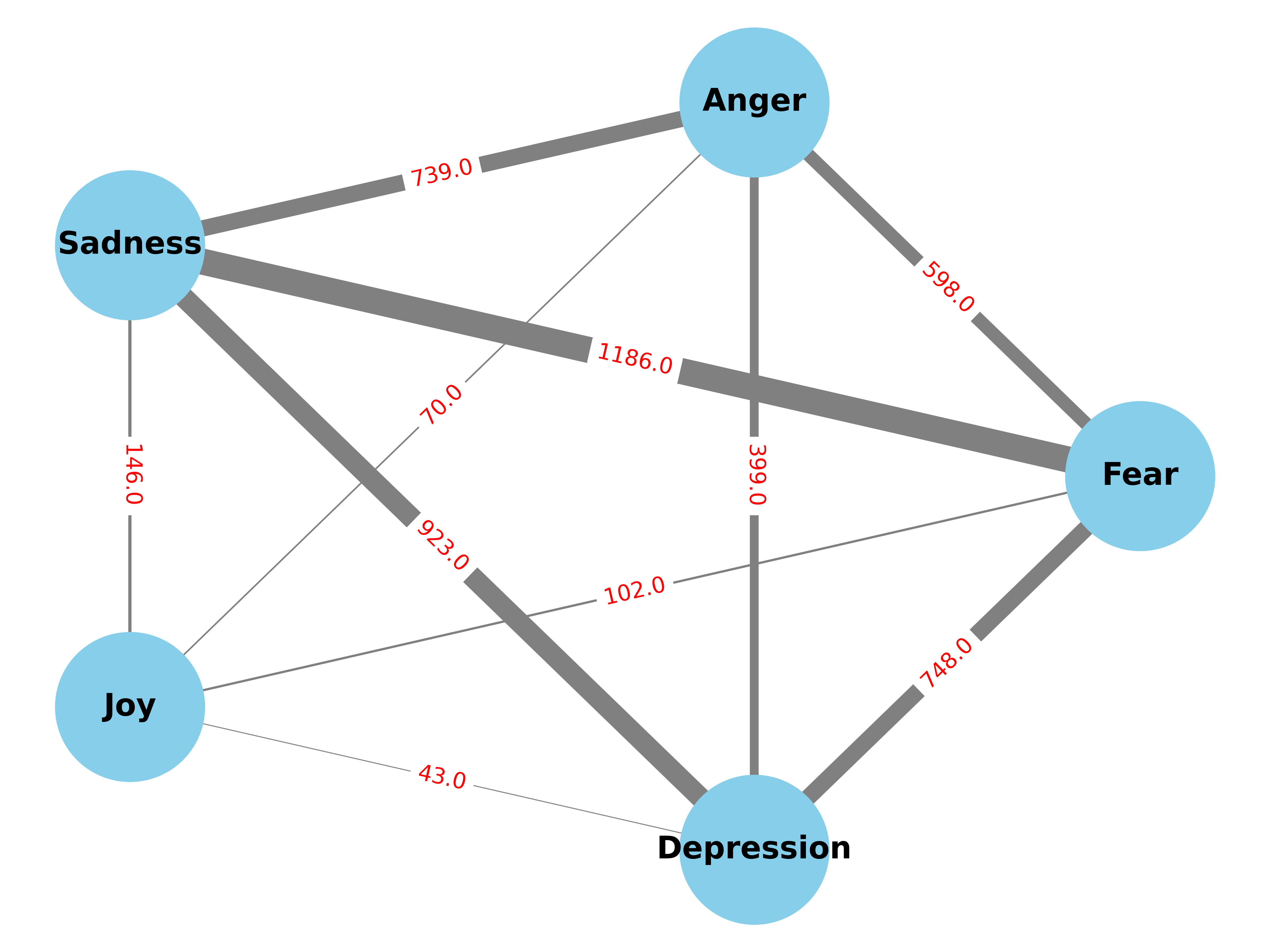} 
    \caption{Emotion co-occurrence network diagram illustrating how different emotions are interconnected within the dataset. Nodes represent emotions, and edges indicate the frequency of their co-occurrence in the same context.}
    \label{fig:emotion_cooccurrence}
\end{figure}

As shown in Figure \ref{fig:emotion_cooccurrence}, ``fear'' and ``sadness'' appeared in most posts with 1,186. This means that an individual who is afraid of something might experience a pure emotion of sadness. However,``sadness'' and ``depression'' appeared in 923 posts. In addition, it is suggested that when a person is ``fear'',``sad'', or ``angry,'' this can lead to mild to severe depression. Similarly, there is also a post that contains both ``joy'' and ``depression'' with 43 of them. Similarly, ``joy'' and ``anger'' share 70 posts together. This analysis is very significant for understanding the emotion shared by multiple labels and is useful in early detection and helping the individual in need. 






\subsubsection{Temporal dynamics of emotions}
\label{emotion_over_time}

Here, Fig. \ref{fig:time_series} shows the overall emotion timeline of NepEMO dataset based on the datetime ranging from January 2019 to mid-2025. It shows that `Sadness' is the highest emotion throughout the years, often staying above 0.4 and even reaching around 0.6 during mid-2023 and 2024. `Fear' also shows a similar trend, with sharp rises and a major peak close to 0.5 in 2020, likely connected to the COVID-19 period. In addition, `Fear' reached its highest score of more than 0.5 around mid-2023. After 2020, both `Fear' and `Sadness' continue to increase at different times, showing people’s ongoing worries and struggles. `Joy', in contrast, stays much lower than `Sadness' and `Fear', mostly ranging between 0.2 and 0.4. However, `Joy' rises clearly during October and November every year, approaching 0.4, which matches the time of great Nepalese festivals such as `Dashain' and `Deepawali'. However, `Depression' is more stable, mostly between 0.3 and 0.4, without very sharp peaks, but still consistently present throughout the time period. `Anger' is the lowest among all emotions, usually staying below 0.3, but shows small, sudden spikes at certain times. In general, negative emotions, sadness, fear, and depression are stronger and more consistent than positive emotions, while joy appears only as short seasonal rises and never crosses the levels of sadness and fear.

\begin{figure}[h!]
    \centering
    \includegraphics[width=\linewidth]{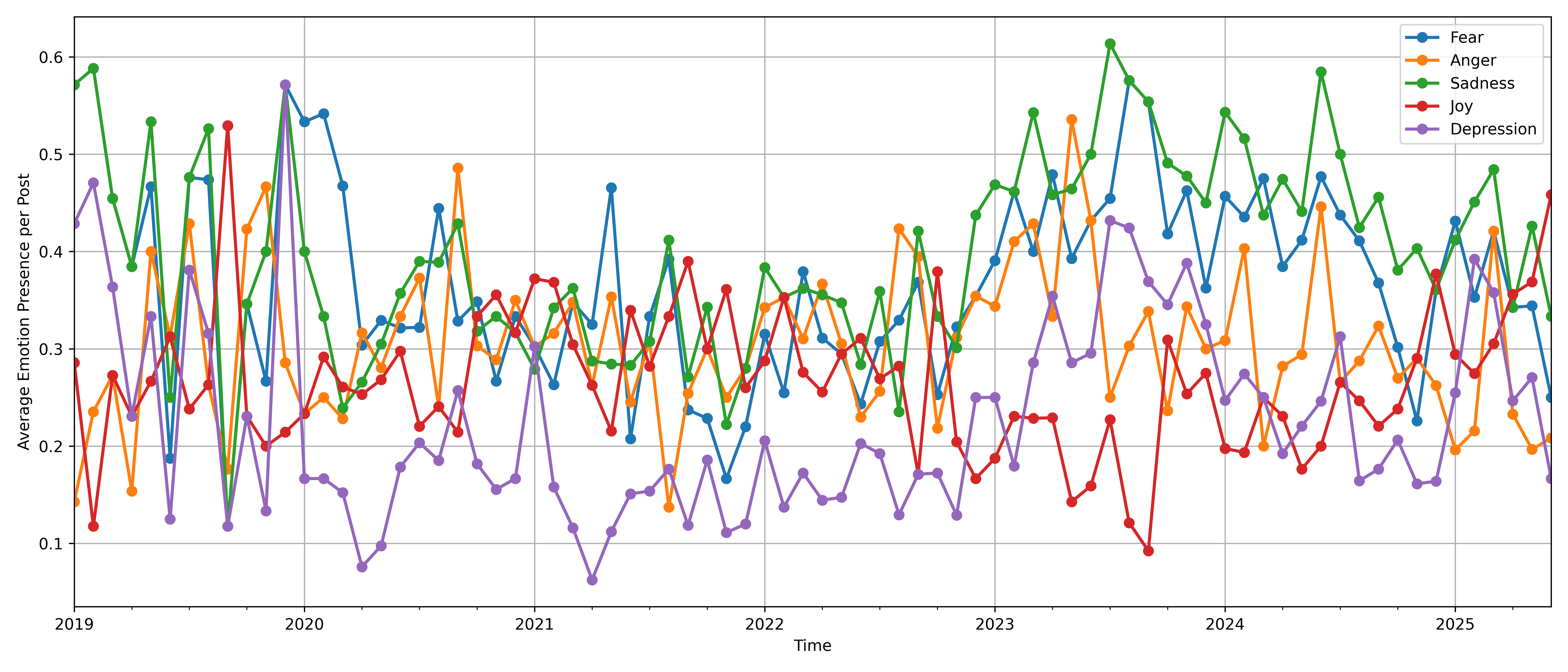} 
    \caption{Emotion over a period of time for a span of six and a half year}
    \label{fig:time_series}
\end{figure}

\section{Baseline Methods}
\label{methods}
In this work, we have used both traditional ML models and DL models to perform MLE and SC tasks. 

\subsection{Traditional ML models}
\label{traditional_ml}
To ensure a strong baseline and compare the performance with the DL models, we implemented widely used ML models for MLE (fear, anger, sadness, joy and depression) and SC (positive, negative and neutral) tasks. The ML models that have been chosen for the experiments are Support Vector Machine (SVM), K-Nearest Neighbour (KNN), Random Forest (RF), Logistic Regression (LR), Decision Tree (DT), and XGBoost \cite{sitoula2025sentiment,joachims2012learning, shah2020comparative, qi2020text}.

\subsection{DL models}
\label{dl_models}
Here, we have employed various models like deep neural network (DNN) \cite{popescu2009multilayer}, CNN \cite{chua1998cnn}, Multi-channel CNN \cite{sitaula2024multi}, LSTM and Bi-LSTM \cite{yu2019review} to extract both local semantic and sequential patterns in text. Additionally, we also experimented with transformer-based models such as XLM-RoBERTa (XLM-R) \cite{barbieri2021xlm}, which can work effectively for multilingual datasets and is also efficient for capturing more complex features, and DistilBERT \cite{cortiz2021exploring}, which is another version of the knowledge-based distillation of the BERT model \cite{koroteev2021bert}. We fine-tuned both models for our dataset to take advantage of contextual embeddings for improved emotion classification and SA \cite{sitoula2025sentiment}. These transformer models are great for handling longer and more complex posts. In addition, it can also work rigorously for MLE classification, as well \cite{chakravarthi2022dravidiancodemix} and can work in various languages and is considered a powerful tool for language processing.

\subsection{Implementation}
\label{Implementation_details}
In this section, we discuss the implementation details of different ML and DL models. We implemented traditional ML models in Scikit-learn \cite{scikit-learn} with optimal hyperparameters for SVM (kernel: linear, class-weight: balanced, probability: True, C: 1.0, random-state: 42), LR (max-iteration: 1000, class-weight: balanced, random-state: 42), KNN (n-neighbors: 5, leaf-size: 30), DT (class-weight: balanced, random-state: 42, max-depth: None, min-samples-split: 2, min-samples-leaf: 1), XGBoost (use-label-encoder: False, eval-metric: logloss, scale-pos-weight: 1, random-state: 42, n-estimators: 100, max-depth: 6, learning-rate: 0.3) and RF (n-estimators: 200, class-weight: balanced, random-state: 42, max-depth: None, min-samples-split: 2, min-samples- leaf: 1) \cite{sitaula2021deep}  

Similarly, for DL models we experimented with different optimizers (like RMSpro and Adam), various learning rates (e.g., $1 \times 10^{-3}$, $1 \times 10^{-4}$, $2 \times 10^{-3}$, $2 \times 10^{-5}$), dropout rates (0.3,0.5).  
The detailed implementation of DL models such as LSTM, Bi-LSTM, DNN, CNN, and MCNN is shown in Table \ref{tab:dl_model_architecture}.

\begin{table}[!h]
 \caption{Parameter details of DL models used in this study. Notation: (64,100, 128) denotes batch size, sequence length and each step with 128 features, respectively; (128) represents the dimension of embedding, k indicates the size of the kernel, 'pool' denotes max-pooling size, Relu indicates Rectified Linear Unit activation and ``-" indicates that the layer is not applicable. The sigmoid is used for binary emotion classification, and softmax is used for sentiment classification.}
\centering
 \renewcommand{\arraystretch}{2} 
 \setlength{\tabcolsep}{10pt} 
 \begin{tabular}{|l|p{2cm}|p{2cm}|p{2cm}|p{2cm}|p{2cm}|}
 \hline
 \textbf{Layer} & \textbf{DNN} & \textbf{CNN} & \textbf{MCNN} & \textbf{LSTM} & \textbf{BiLSTM} \\
 \hline
Input & (64,100) & (64,100,128) & (64,100,128) & (64,100,128) & (64,100,128) \\
\hline
Embedding & (128) & (128) & (128) & (128) & (128) \\
\hline
 Conv1D & — & (128, k=3, Relu) & ( 128, k=3,4,5, Relu) & — & — \\
 \hline
 Pooling & — & pool=2 & pool=2 per filter & — & — \\
 \hline
 Dense & (128, 64, Relu) & (128, Relu) & (128, Relu) & — & — \\
 \hline
 Dropout & 0.4, 0.3 & 0.3 & 0.3 & 0.3 & 0.3 \\
 \hline
 sigmoid  & 2 & 2 & 2 & 2 &2 \\
 \hline
softmax  & 3 & 3 & 3 & 3 & 3 \\
 \hline
 \end{tabular}
 \label{tab:dl_model_architecture}
 \end{table}

Additionally, for transformer-based models such as XLM-R and DistilBERT, we fine-tuned for both binary emotion label classification and three-class sentiment. Using pre-trained embeddings and learning rates of $1 \times 10^{-5}$ to $5 \times 10^{-5}$, batch size ranging from 16-32 and early stopping of 3 is used to achieve the best performance. All DL models were trained using the NVIDIA Tesla P100 GPU. 

\subsection{Performance metrics}
\label{performance_metrics}
To evaluate the performance of the models used in our study, we used Precision (P) (Eq.\ref{eq:precision}), Recall (R) (Eq.\ref{eq:recall}), F1 score (F) (Eq.\ref{eq:f1}), Accuracy (A) (Eq.\ref{eq:accuracy}), and AUC (A'). These metrics are widely used in ML and NLP for classification tasks \cite{powers2020evaluation, scikit-learn}. For binary class and multi-class tasks, we report macro-averaged values. The AUC score for binary classification was calculated using the predicted probability of yes or no labels (Eq\ref{eq:binary_auc}). Similarly, for the multi-class SA task, the AUC score was calculated using the one-vs-rest (OVR) strategy in Eq.\ref{eq:macro_auc} \cite{hand2001simple}. 

{\small
\begin{equation}
\text{P} = \frac{TP}{TP + FP}
\label{eq:precision}
\end{equation}
}

{\small
\begin{equation}
\text{R} = \frac{TP}{TP + FN}
\label{eq:recall}
\end{equation}
}

{\small
\begin{equation}
\text{F} = 2 \cdot \frac{\text{Precision} \cdot \text{Recall}}{\text{Precision} + \text{Recall}}
\label{eq:f1}
\end{equation}
}

{\small
\begin{equation}
\text{A} = \frac{TP + TN}{TP + TN + FP + FN}
\label{eq:accuracy}
\end{equation}
}

{\small
\begin{equation}
\text{True Positive Rate (TPR)} = \frac{TP}{TP + FN}
\label{eq:binary_tpr}
\end{equation}
}

{\small
\begin{equation}
\text{False Positive Rate (FPR)} = \frac{FP}{FP + TN}
\label{eq:binary_fpr}
\end{equation}
}

{\small
\begin{equation}
\text{A'}_{binary} = \int_0^1 \text{TPR}(\text{FPR}) \, d(\text{FPR})
\label{eq:binary_auc}
\end{equation}
}

{\small
\begin{equation}
\text{A'}_{macro} = \frac{1}{K} \sum_{k=1}^{K} \text{AUC}_k
\label{eq:macro_auc}
\end{equation}
}

{\small
\begin{multline}
\text{A'}_k 
\text{ is computed treating class $k$ as } \\
\text{positive and all other classes as negative One-vs-Rest, OvR).}
\label{eq:auc_ovr}
\end{multline}
}

\subsection{Classification}

Due to the nature of their labels, we have used \textbf{emotion classification} (fear, sadness, anger, joy, depression) and \textbf{sentiment classification} (positive, negative, neutral) differently in our task. Sentiment is mutually exclusive, such that each post can belong to only one sentiment category, while a specific post can have multiple emotions, and they co-exist. Therefore, we model emotions as independent binary classification tasks. 

\subsubsection{Emotion classification}
We have used the sigmoid activation function in the output layer for each emotion label. For each emotion, we design a separate classifier that predicts whether the emotion is present (label = 1) or absent (label = 0). The sigmoid activation function is used in the output layer to map the raw score $z$ to a probability between $0$ and $1$:

{\small
\begin{equation}
\sigma(z) = \frac{1}{1 + e^{-z}}
\label{eq:sigmoid}
\end{equation}
}

The final prediction is obtained by applying a threshold (for example, 0.5) on $\sigma(z)$.

The loss function used is the \textbf{Binary Cross-Entropy (BCE)} loss, defined in Eq.\ref{eq:bce}:

{\small
\begin{equation}
\mathcal{L}_{BCE} = - \frac{1}{N} \sum_{i=1}^{N} 
\left[ y_i \log(\hat{y}_i) + (1 - y_i) \log(1 - \hat{y}_i) \right]
\label{eq:bce}
\end{equation}
}

{\small
where $y_i \in \{0,1\}$ is the true label for a given emotion, $\hat{y}_i$ is the predicted probability, 
and $N$ is the total number of samples. This loss is applied separately for each of the five emotion tasks.
}

\subsubsection{Sentiment classification}
We have applied the softmax activation function for sentiment classification for three output classes: positive, negative, and neutral. The softmax activation function ensures that the predicted probabilities sum to 1. The softmax is shown in Eq.\ref{eq:softmax}

{\small
\begin{equation}
P(y = j \mid \mathbf{z}) = \frac{e^{z_j}}{\sum_{k=1}^{3} e^{z_k}}, \quad j \in \{1,2,3\}
\label{eq:softmax}
\end{equation}
}

The objective is to minimise CCE loss during training, which helps the model to allocate the highest probability to the correct sentiment class. Therefore, we apply \textbf{Categorical Cross-Entropy (CCE)} for multi class sentiment classification as shown in Eq.\ref{eq:cce}

{\small
\begin{equation}
\mathcal{L}_{CCE} = - \frac{1}{N} \sum_{i=1}^{N} \sum_{j=1}^{3} 
y_{ij} \log(\hat{y}_{ij})
\label{eq:cce}
\end{equation}
}

{\small
where $y_{ij}$ is the one-hot encoded true label for sentiment class $j$ and $\hat{y}_{ij}$ is the 
predicted probability. 
}
\section{Results and discussion}
\label{result_and_discussion}
Here, we present results obtained from all models under consideration for the emotion and sentiment classification. 

\subsection{Performance on Fear and Sadness}
Table \ref{tab:fear} and Table \ref{tab:sadness} show the results for the detection of emotions of fear and sadness, respectively. Among all traditional ML models, the best performing models are LR and RF with test accuracy of 77.83\% and 78.16\%, respectively, for fear and 77.94\% and 77.72\%, respectively, for sadness. 
In contrast, KNN and DT are two models that achieved a weaker score with accuracy of 71.33\% and 64.73\% for fear and 71.11\% and 66.63\% for sadness, respectively. Regarding the performance of DL models, CNN, LSTM and DNN provide competitive results, but are significantly behind ML models. However, transformer-based models such as XLM-R and DistilBERT achieve the best results among all models with an accuracy of 78.39\% and 76.60\% for fear in terms of XLM-R and DistilBERT, respectively. Similarly, for sandess, XLM-R has the best accuracy score of 78.39\% and DistilBERT has an accuracy of 80.29\%.

For example, fear was often expressed with phrases such as \textit{``dar lagcha''} (I feel scared), or in a long context like \textit{``entrance ni pass hudina jasto laagcha''} (I feel like I cannot pass the entrance). However, sadness appeared in direct emotions like \textit{``man dherai dukheko cha''} (the heart feels very hurt) or \textit{``Uhako wife ko nidhan vaye dekhi''} (since his wife passed away). These important phrases were highly beneficial for our model in detecting and predicting these emotions.

\begin{table}[!h]
\centering
\renewcommand{\arraystretch}{1.3}
\setlength{\tabcolsep}{6pt}
\caption{Performance metrics of ML and DL models for Fear. 
P, R, F, A, and A' denote Precision, Recall, F1-score, Accuracy, and AUC score, respectively. 
Bold indicates the highest score for each metric.}
\begin{tabular}{l|c|c|c|c|c}
\toprule
\multicolumn{6}{c}{\textbf{Fear}} \\
\cmidrule(lr){1-6}
\textbf{Model} & \textbf{P} & \textbf{R} & \textbf{F} & \textbf{A} & \textbf{A'} \\
\midrule
SVM & 0.7338 & 0.7451 & 0.7374 & 0.7503 & 0.8114 \\
LR & 0.7617 & 0.7717 & 0.7654 & 0.7783 & 0.8354 \\
KNN & 0.7117 & 0.6377 & 0.6400 & 0.7133 & 0.7318 \\
DT & 0.6273 & 0.6332 & 0.6286 & 0.6473 & 0.6328 \\
XGBoost & 0.7530 & 0.7300 & 0.7377 & 0.7671 & 0.7671 \\
RF & \textbf{0.7739} & 0.7408 & 0.7508 & 0.7816 & 0.8306 \\ \hline
LSTM & 0.7289 & 0.7341 & 0.7311 & 0.7480 & 0.8000 \\
Bi-LSTM & 0.7166 & 0.7310 & 0.7187 & 0.7290 & 0.8046 \\
CNN & 0.7532 & 0.7335 & 0.7404 & 0.7682 & 0.8163 \\
MCNN & 0.7348 & 0.6977 & 0.7066 & 0.7469 & 0.8009 \\
DNN & 0.7570 & 0.7417 & 0.7475 & 0.7727 & 0.8286 \\ \hline\hline
DistilBERT & 0.7528 & 0.7686 & 0.7563 & 0.7660 & 0.8482 \\
XLM-R & 0.7676 & \textbf{0.7781} & \textbf{0.7715} & \textbf{0.7839} & \textbf{0.8607} \\
\bottomrule
\end{tabular}
\label{tab:fear}
\end{table}

\begin{table}[!h]
\centering
\renewcommand{\arraystretch}{1.3}
\setlength{\tabcolsep}{6pt}
\caption{Performance metrics of ML and DL models for Sadness. 
P, R, F, A, and A' denote Precision, Recall, F1-score, Accuracy, and AUC score, respectively. 
Bold indicates the highest score for each metric.}
\begin{tabular}{l|c|c|c|c|c}
\toprule
\multicolumn{6}{c}{\textbf{Sadness}} \\
\cmidrule(lr){1-6}
\textbf{Model} & \textbf{P} & \textbf{R} & \textbf{F} & \textbf{A} & \textbf{A'} \\
\midrule
SVM & 0.7662 & 0.7642 & 0.7651 & 0.7772 & 0.8466 \\
LR & 0.7683 & 0.7691 & 0.7687 & 0.7794 & 0.8536 \\
KNN & 0.7334 & 0.6504 & 0.6492 & 0.7111 & 0.7528 \\
DT & 0.6507 & 0.6522 & 0.6513 & 0.6663 & 0.6522 \\
XGBoost & 0.7634 & 0.7488 & 0.7539 & 0.7716 & 0.8302 \\
RF & 0.7715 & 0.7519 & 0.7582 & 0.7772 & 0.8349 \\ \hline
LSTM & 0.7594 & 0.7675 & 0.7620 & 0.7693 & 0.8423 \\
Bi-LSTM & 0.7274 & 0.7145 & 0.7188 & 0.7391 & 0.7816 \\
CNN & 0.7460 & 0.7465 & 0.7463 & 0.7581 & 0.8233 \\
MCNN & 0.7585 & 0.7706 & 0.7592 & 0.7637 & 0.8353 \\
DNN & 0.7573 & 0.7427 & 0.7477 & 0.7660 & 0.8367 \\ \hline\hline
DistilBERT & \textbf{0.7975} & 0.7828 & \textbf{0.7882} & \textbf{0.8029} & 0.8787 \\
XLM-R & 0.7802 & \textbf{0.7839} & 0.7803 & 0.7839 & \textbf{0.8792} \\
\bottomrule
\end{tabular}
\label{tab:sadness}
\end{table}

\subsection{Performance on Anger and Joy}

From Table \ref{tab:anger} and Table \ref{tab:joy}, we can clearly see that all our models have achieved a significant result in anger detection compared to joy prediction. Traditional ML and DL models have achieved high accuracy for anger post-prediction ranging from 69.65\% to 81.30\%. In addition, transformer-based methods such as DistilBERT and XLM-R achieved the highest accuracy score among all with 83.43\% and 86.00\%, respectively. However, our models has shown a slightly lower score for joy emotion detection with the best accuracy of 82.75\% achieved by the XLM-R model. However, compared to the DT model, which has an accuracy of 69.65\% for fear, a better result was achieved with 71.22\% for joy on the same metric. The main reason for this can be that anger is more directly expressed, while joy tends to be shown in a more diverse manner, making it harder for models to capture it. 

For example, anger expression includes phrases like \textit{``yo system le dherai ris uthako cha''} (this system makes me very angry), while joy can be like \textit{``khushi lagcha saathi haru sanga bhetera''} (I fell happy to meet friends). 

\begin{table}[!h]
\centering
\renewcommand{\arraystretch}{1.3}
\setlength{\tabcolsep}{6pt}
\caption{Performance metrics of ML and DL models for Anger. 
P, R, F, A, and A' denote Precision, Recall, F1-score, Accuracy, and AUC score, respectively. 
Bold indicates the highest score for each metric.}
\begin{tabular}{l|c|c|c|c|c}
\toprule
\multicolumn{6}{c}{\textbf{Anger}} \\
\cmidrule(lr){1-6}
\textbf{Model} & \textbf{P} & \textbf{R} & \textbf{F} & \textbf{A} & \textbf{A'} \\
\midrule
SVM & 0.7591 & 0.7761 & 0.7659 & 0.7940 & 0.8555 \\
LR & 0.7806 & 0.8020 & 0.7888 & 0.8130 & 0.8704 \\
KNN & 0.7152 & 0.6076 & 0.6104 & 0.7380 & 0.7380 \\
DT & 0.6514 & 0.6611 & 0.6549 & 0.6965 & 0.6611 \\
XGBoost & 0.7863 & 0.7429 & 0.7580 & 0.8085 & 0.8458 \\
RF & 0.8102 & 0.7298 & 0.7516 & 0.8130 & 0.8548 \\ \hline
LSTM & 0.7468 & 0.7174 & 0.7282 & 0.7816 & 0.7816 \\
Bi-LSTM & 0.7303 & 0.7134 & 0.7203 & 0.7704 & 0.8065 \\
CNN & 0.7398 & 0.7455 & 0.7425 & 0.7783 & 0.8388 \\
MCNN & 0.7205 & 0.7526 & 0.7253 & 0.7458 & 0.8405 \\
DNN & 0.7780 & 0.7661 & 0.7714 & 0.8096 & 0.8620 \\ \hline\hline
DistilBERT & 0.8036 & 0.8225 & 0.8114 & 0.8343 & 0.9045 \\
XLM-R & \textbf{0.8353} & \textbf{0.8360} & \textbf{0.8356} & \textbf{0.8600} & \textbf{0.9207} \\
\bottomrule
\end{tabular}
\label{tab:anger}
\end{table}

\begin{table}[!h]
\centering
\renewcommand{\arraystretch}{1.3}
\setlength{\tabcolsep}{6pt}
\caption{Performance metrics of ML and DL models for Joy. 
P, R, F, A, and A' denote Precision, Recall, F1-score, Accuracy, and AUC score, respectively. 
Bold indicates the highest score for each metric.}
\begin{tabular}{l|c|c|c|c|c}
\toprule
\multicolumn{6}{c}{\textbf{Joy}} \\
\cmidrule(lr){1-6}
\textbf{Model} & \textbf{P} & \textbf{R} & \textbf{F} & \textbf{A} & \textbf{A'} \\
\midrule
SVM & 0.7022 & 0.7171 & 0.7084 & 0.7626 & 0.7886 \\
LR & 0.7179 & 0.7284 & 0.7226 & 0.7772 & 0.7976 \\
KNN & 0.7260 & 0.6939 & 0.7059 & 0.7850 & 0.7420 \\
DT & 0.6439 & 0.6562 & 0.6486 & 0.7122 & 0.6566 \\
XGBoost & 0.7538 & 0.6721 & 0.6925 & 0.7940 & 0.7842 \\
RF & 0.7788 & 0.6619 & 0.6843 & 0.7984 & 0.7702 \\ \hline
LSTM & 0.6965 & 0.7222 & 0.7051 & 0.7525 & 0.7713 \\
Bi-LSTM & 0.7287 & 0.6806 & 0.6961 & 0.7850 & 0.7748 \\
CNN & 0.6797 & 0.7011 & 0.6871 & 0.7391 & 0.7655 \\
MCNN & 0.7142 & 0.7130 & 0.7136 & 0.7760 & 0.7647 \\
DNN & 0.7549 & 0.5661 & 0.5537 & 0.7592 & 0.7497 \\ \hline\hline
DistilBERT & 0.7588 & 0.7553 & 0.7570 & 0.8108 & 0.8530 \\
XLM-R & \textbf{0.7816} & \textbf{0.7708} & \textbf{0.7758} & \textbf{0.8275} & \textbf{0.8678} \\
\bottomrule
\end{tabular}
\label{tab:joy}
\end{table}

\subsection{Performance on Depression and Sentiment}
We present the results of the depression and sentiment classification, respectively, in Tables \ref{tab:depression} and  \ref{tab:sentiment}. For depression detection, the RF model shows reasonably strong results, achieving the highest precision of 80.93\% among all, but the lower recall with 63.86\%. XLM-R achieve the best balance with an F1-score of 79.91\%, accuracy of 86.34\% and an AUC score of 88.55\%. This result shows that transformer-based models are very good and capable of detecting depression-related text much better than traditional and DL models. Meanwhile, sentiment classification is more challenging, and results are low compared to binary emotion tasks. Traditional ML models, such as SVM and LR, perform moderately well, whereas DL models, such as CNN and the Bi-LSTM model, achieve similar results. The transformer-based models again show superior results with DistilBERT achieving the best precision (68.88\%) and the XLM-R model achieving the highest recall, F1-score and AUC with 64.34\%, 65.63\% and 85.65\%, respectively. 
These results show that compared to depression detection, sentiment detection, such as positive, negative, and neutral, is much challenging on Nepali SM posts, due to its mixed expression and code change language. For example, depression-related expressions include phrases like \textit{``jindagi ma kunai aasha chaina''} (there is no hope in life), {``malai khali sucide garna maan lagcha''} (I always feel like doing suicide). However, in terms of general sentiment, positive tones appeared in phrases such as \textit{``yo din dherai ramailo cha''} (this day is very joyful), while negative tones appeared in \textit{``yo neta haru sabai chor ho''} (All these politicians are thieves). Neutral sentiment phrases include such as \textit{``Please kehi suggestion dinus na''} (Please give some suggestions).

\begin{table}[!h]
\centering
\renewcommand{\arraystretch}{1.3}
\setlength{\tabcolsep}{6pt}
\caption{Performance metrics of ML and DL models for Depression. 
P, R, F, A, and A' denote Precision, Recall, F1-score, Accuracy, and AUC score, respectively. 
Bold indicates the highest score for each metric.}
\begin{tabular}{l|c|c|c|c|c}
\toprule
\multicolumn{6}{c}{\textbf{Depression}} \\
\cmidrule(lr){1-6}
\textbf{Model} & \textbf{P} & \textbf{R} & \textbf{F} & \textbf{A} & \textbf{A'} \\
\midrule
SVM & 0.7240 & 0.7600 & 0.7380 & 0.8096 & 0.8509 \\
LR & 0.7411 & 0.7851 & 0.7576 & 0.8219 & 0.8647 \\
KNN & 0.7854 & 0.6514 & 0.6805 & 0.8320 & 0.7550 \\
DT & 0.6193 & 0.6281 & 0.6231 & 0.7380 & 0.6281 \\
XGBoost & 0.7250 & 0.6817 & 0.6980 & 0.8163 & 0.8242 \\
RF & \textbf{0.8093} & 0.6386 & 0.6678 & 0.8331 & 0.8402 \\ \hline
LSTM & 0.7054 & 0.7109 & 0.7081 & 0.8018 & 0.7997 \\
Bi-LSTM & 0.7025 & 0.7025 & 0.7025 & 0.8007 & 0.7858 \\
CNN & 0.7102 & 0.7219 & 0.7156 & 0.8040 & 0.8039 \\
MCNN & 0.7478 & 0.7133 & 0.7276 & 0.8298 & 0.7572 \\
DNN & 0.7770 & 0.7096 & 0.7335 & 0.8421 & 0.8553 \\ \hline\hline
DistilBERT & 0.7952 & 0.7450 & 0.7652 & 0.8555 & 0.8733 \\
XLM-R & 0.7949 & \textbf{0.8038} & \textbf{0.7991} & \textbf{0.8634} & \textbf{0.8855} \\
\bottomrule
\end{tabular}
\label{tab:depression}
\end{table}

\begin{table}[!h]
\centering
\renewcommand{\arraystretch}{1.3}
\setlength{\tabcolsep}{6pt}
\caption{Performance metrics of ML and DL models for Sentiment. 
P, R, F, A, and A' denote Precision, Recall, F1-score, Accuracy, and AUC score, respectively. 
Bold indicates the highest score for each metric.}
\begin{tabular}{l|c|c|c|c|c}
\toprule
\multicolumn{6}{c}{\textbf{Sentiment}} \\
\cmidrule(lr){1-6}
\textbf{Model} & \textbf{P} & \textbf{R} & \textbf{F} & \textbf{A} & \textbf{A'} \\
\midrule
SVM & 0.6094 & 0.6034 & 0.6048 & 0.6282 & 0.7941 \\
LR & 0.6308 & 0.6240 & 0.6260 & 0.6461 & 0.7942 \\
KNN & 0.5669 & 0.5206 & 0.5058 & 0.5095 & 0.7151 \\
DT & 0.4712 & 0.4710 & 0.4702 & 0.4950 & 0.6051 \\
XGBoost & 0.6251 & 0.5802 & 0.5926 & 0.6327 & 0.7680 \\
RF & 0.6503 & 0.5888 & 0.6031 & 0.6405 & 0.7775 \\ \hline
LSTM & 0.5923 & 0.6044 & 0.5928 & 0.6047 & 0.7748 \\
Bi-LSTM & 0.5947 & 0.5950 & 0.5943 & 0.6204 & 0.7857 \\
CNN & 0.5531 & 0.5670 & 0.5474 & 0.5812 & 0.7719 \\
MCNN & 0.6290 & 0.6141 & 0.6162 & 0.6349 & 0.7884 \\
DNN & 0.5667 & 0.4808 & 0.4773 & 0.5801 & 0.7572 \\ \hline\hline
DistilBERT & \textbf{0.6888} & 0.6328 & 0.6483 & 0.6831 & 0.8302 \\
XLM-R & 0.6849 & \textbf{0.6434} & \textbf{0.6563} & \textbf{0.6876} & \textbf{0.8565} \\
\bottomrule
\end{tabular}
\label{tab:sentiment}
\end{table}


\section{Conclusion and future works}
\label{conclusion}
In this study, we presented a Nepali Reddit dataset annotated for multi-emotional and sentimental classification. We evaluated the performance of various traditional ML, DL and transformer-based models, where transformer-based methods like XLM-R and DistilBERT showed superiority compared to ML and DL models across all emotion and sentiment tasks. Transformer-based models were able to capture complex semantic and code-mixed linguistic patterns efficiently.
Beyond classification, our exploratory studies provided valuable insights into NepEMO. TF-IDF-based feature extraction identified emotion and sentiment-specific keywords, whereas n-gram analysis identified multi-word expressions indicating common emotional or sentiment themes. Similarly, network diagrams showed co-occurrence links between key phrases, showing links between emotions and mental health issues. Topic modelling with LDA, validated by visualisation and coherence scores, revealed latent topic patterns and improved interpretability.

Importantly, a temporal analysis from 2019 to 2025 identified patterns in emotion dynamics over time. By evaluating changes quarterly, we were able to determine which feelings and sentiments were increasing or decreasing, offering significant information on the growing social and mental health trends of the Nepali online community. In general, we discovered that the combination of advanced modelling techniques, feature extraction, visualisation tools, and temporal analysis can effectively capture the complexity of multilingual code-mixed social media content. 

This work provides a solid platform for future research in mental health monitoring, sentiment tracking, and subtle emotion identification in low-resource languages, while offering practical insights for computational modelling and social media analysis.
However, it has two main limitations. First, the current dataset only includes five emotion categories with 4,462 posts, which may not provide sufficient information or represent the overall scenario of emotion. Second, the baseline models have achieved lower classification accuracy. In future, a larger dataset with additional categories such as stress and hate could contribute more to emotion detection. Furthermore, a lightweight classification model can be developed that can deliver much better results with less computational power than multilingual transformer models.

\backmatter








\section*{Declarations}


\subsection*{Funding}
This research receive no external funding.
\subsection*{Conflict of interest}
Authors declare no conflict of interest.
\subsection*{Data availability}
The data will be made available at the following GitHub repository: https://github.com/Sameer67/Nepali-Reddit-NepEMO-/tree/main.
\subsection*{ Author contribution }
SS, TBS: Conceptualisation, Methodology, Software, Writing – Original draft, Visualisation, Investigation, Validation. SS, LPB, AP: Data curation, TBS, SS, AN: Writing – Review \& Editing.







\begin{appendices}






\end{appendices}


\bibliography{sn-bibliography}

\end{document}